\documentclass{article}

\PassOptionsToPackage{numbers, compress}{natbib}

\usepackage[preprint]{neurips_2025}

\usepackage[utf8]{inputenc} % allow utf-8 input
\usepackage[T1]{fontenc}    % use 8-bit T1 fonts
\usepackage{hyperref}       % hyperlinks
\usepackage{url}            % simple URL typesetting
\usepackage{booktabs}       % professional-quality tables
\usepackage{amsfonts}       % blackboard math symbols
\usepackage{nicefrac}       % compact symbols for 1/2, etc.
\usepackage{microtype}      % microtypography
\usepackage{xcolor}         % colors
\usepackage{colortbl}  %彩色表格需要加载的宏包
\usepackage{graphicx} 
\usepackage{subcaption}  % 使用这个替代 subfigure
\usepackage{amsmath}
\usepackage{algorithm}
\usepackage[noend]{algpseudocode}
\usepackage{multirow}
\usepackage{makecell}
\usepackage{wrapfig}
\usepackage{tcolorbox}

\usepackage{tikz}
\usetikzlibrary{arrows.meta}

\newcommand{\eqrefe}[1]{Eq.~\eqref{#1}}
\newcommand{\figrefe}[1]{Fig.~\ref{#1}}
\newcommand{\tabrefe}[1]{Tab.~\ref{#1}}

\newcommand{\increase}[1]{\tiny\textcolor[HTML]{00A86B}{$\uparrow${#1}}}
\newcommand{\decrease}[1]{\tiny\textcolor[HTML]{E34234}{$\downarrow${#1}}}
\renewcommand{\paragraph}[1]{\vspace{0.05in}\noindent\textbf{#1.}\,\,}

\title{Towards Self-Improvement of Diffusion Models \\ via Group Preference Optimization}

\author{Renjie Chen, Wenfeng LIN, Yichen Zhang, Jiangchuan Wei, Boyuan Liu, \\ \bf Chao Feng, Jiao Ran, Mingyu Guo\\
    ByteDance Douyin Content Group \\
    \texttt{\{chenrenjie.1998, linwenfeng.1008,zhangyichen.99,weijiangchuan\}} \\
    \texttt{\{liuboyuan,chaofeng.zz, ranjiao,guomingyu.313\}@bytedance.com}
}

\begin{document}

\maketitle

\begin{abstract}
% https://arxiv.org/pdf/2410.18013
% Moreover, unlike text, which can be easily revised, these difficult-to-edit images quickly become outdated as advances in T2I models produce higher-quality data. 
Aligning text-to-image (T2I) diffusion models with Direct Preference Optimization (DPO) has shown notable improvements in generation quality. 
However, applying DPO to T2I faces two challenges: the sensitivity of DPO to preference pairs and the labor-intensive process of collecting and annotating high-quality data. 
In this work, we demonstrate that preference pairs with marginal differences can degrade DPO performance. Since DPO relies exclusively on relative ranking while disregarding the absolute difference of pairs, it may misclassify losing samples as wins, or vice versa. We empirically show that extending the DPO from pairwise to groupwise and incorporating reward standardization for reweighting leads to performance gains without explicit data selection. 
Furthermore, we propose \textbf{Group Preference Optimization (GPO)}, an effective self-improvement method that enhances performance by leveraging the model's own capabilities without requiring external data. 
Extensive experiments demonstrate that GPO is effective across various diffusion models and tasks. Specifically, combining with widely used computer vision models, such as YOLO and OCR, the GPO improves the accurate counting and text rendering capabilities of the \textit{Stable Diffusion 3.5 Medium} by 20 percentage points. Notably, as a plug-and-play method, no extra overhead is introduced during inference. 

\end{abstract}

\section{Introduction}
Text-to-image diffusion models\cite{rombach2022high,podellsdxl,esser2024scaling,betker2023improving} pretrained on large-scale internet datasets\cite{schuhmann2022laion,schuhmann2021laion} exhibit remarkable capabilities in generating high-quality and creative images from textual prompts. However, even state-of-the-art T2I models still suffer from several well-known limitations, including poor prompt understanding \cite{chatterjee2024getting}, inaccurate object counting\cite{binyamin2024make,cao2025text}, and difficulty in rendering legible text\cite{liu2022character,tuo2023anytext,chen2023textdiffuser}. Several approaches attempt to mitigate these issues, such as scaling up the capacity of the diffusion model\cite{esser2024scaling, flux2024}, using detailed captions \cite{betker2023improving}, or improving text encoders \cite{esser2024scaling,ma2024exploring}. These methods require models to be trained from scratch, making them difficult to adapt to existing models. An alternative approach involves introducing additional conditions \cite{zhang2023adding,binyamin2024make,tuo2023anytext,chen2023textdiffuser} to the pre-trained model, but increasing the complexity of the generation pipeline.

Inspired by the success of \textit{reinforcement learning from human feedback (RLHF)} in Large Language Models (LLMs), training a reward model to align human preference, and fine-tuning T2I diffusion models with RL algorithms shows promise to alleviate the limitations of the diffusion model. Nevertheless, backpropagation through the diffusion trajectories requires a differentiable reward model and significant memory, which limits the scalability to large diffusion models. Therefore, Diff-DPO\cite{liu2024alignment} and its variants\cite{yang2024using, liang2025aestheticposttrainingdiffusionmodels, zhang2025diffusion} apply \textit{direct preference optimization (DPO)} \cite{lee2023aligning} to diffusion, eliminating the need for an explicit reward model and training on human-annotated preference pairs directly.

\begin{figure}[htbp]
    \centering
    \includegraphics[width=\textwidth]{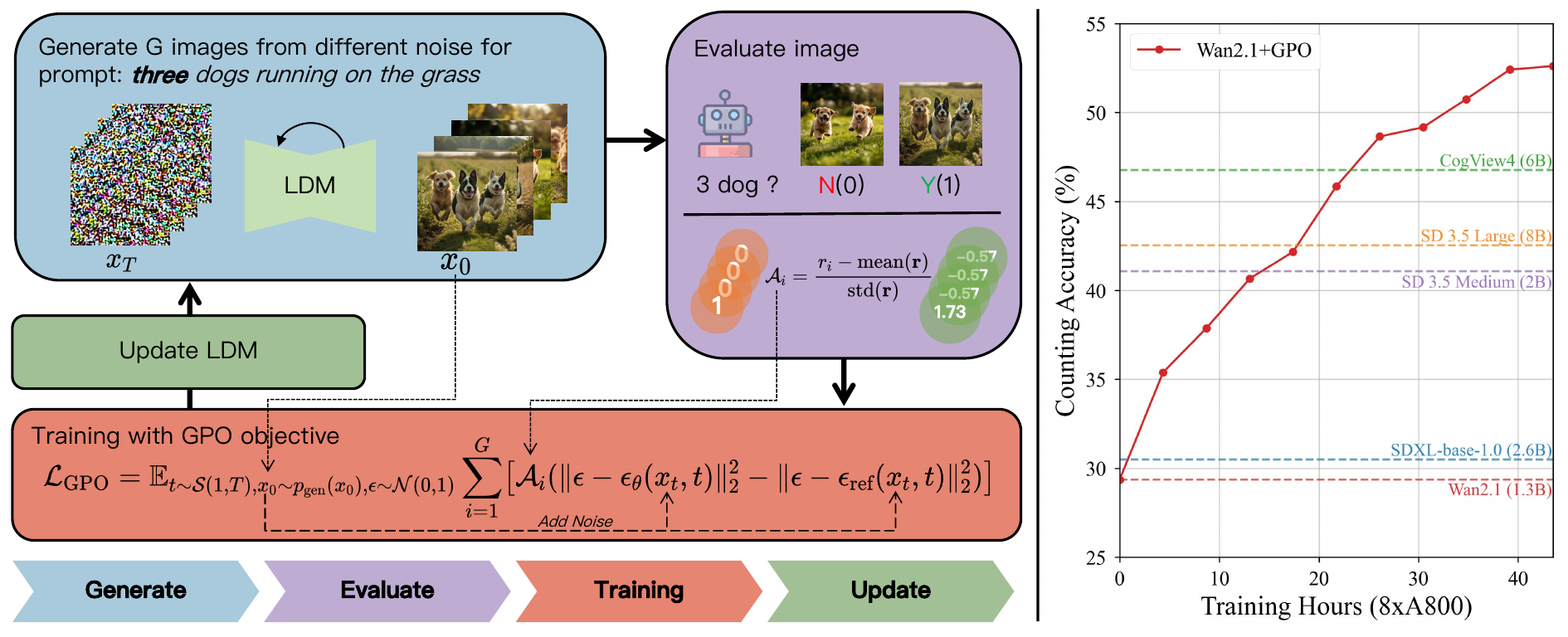}
    \caption{Overview of Group Preference Optimization. Combined with YOLO v11, our approach enables a 1.3B parameter model to surpass larger state-of-the-art models in accurate counting tasks.}
    \label{fig:teaser}
\end{figure}

Unfortunately, the application of DPO in T2I faces two challenges: Firstly, DPO is sensitive to data quality\cite{li2023policy, deng2025less} and may even experience a significant performance degradation compared to RLHF. Secondly, high-quality preference data is labor-intensive, especially for the image. For example, Pick-a-Pic\cite{kirstain2023pick} collects images generated by existing generative models, followed by human annotation to obtain pairwise preference. The entire collection process cost ~\$50K \cite{karthik2024scalable}. 
Moreover, unlike the text, which can be directly modified, difficult-to-edit images will become outdated as the rapid advancement of T2I models produces higher-quality images.

In this paper, we aim to alleviate the limitations mentioned earlier and achieve self-improvement of the diffusion model without external human-annotated datasets. 
First of all, we reveal that the preference pair margin, i.e., the magnitude by which the winning sample outperforms the losing one, significantly impacts DPO performance. Filtering out preference pairs with negligible margin improves DPO performance, as relative ranking alone fails to account for absolute quality differences and mistakenly classifies the losing samples as wins or vice versa. 
Previous works\cite{lee2025calibrated,liu2024videodpo,deng2025less,karthik2024scalable} propose to use pair filtering or reward calibration to alleviate the influence of pair margin, yet remain confined to pairwise comparisons. Differently, we extend DPO from pairwise to groupwise and introduce a reward standardization method that reassigns coefficients to samples without requiring data selection. 
Specifically, given a prompt and the corresponding $G$ images, the sample score is calculated using model-based or rule-based metrics. We establish the group baseline by taking the average score of all samples. Following the DPO paradigm, we set a goal to encourage the model to increase the generation probability of samples that exceed baseline, while suppressing samples that are below baseline. For stable optimization, we adopt standardized scores (i.e., z-scores) as the weighting coefficient, which has a dual purpose: 1) to provide the relative preference signal within the group, and 2) to ensure the training stability through variance normalization.

Furthermore, the current SOTA models have demonstrated the potential to generate images that align with prompts, but the generation is unstable. Leveraging this property, we propose \textbf{Group Preference Optimization (GPO)}, an effective approach that uses reward standardization training on a group of online-generated data from the model itself. As illustrated in \figrefe{fig:teaser}, the training of GPO does not require the introduction of any external data. When combined with YOLO, GPO can enable the 1.3B Wan2.1 \cite{wang2025wan} model to outperform 6B CogView4 \cite{zheng2024cogview3} in terms of accurate counting ability. The main contributions can be summarized as follows:
\begin{itemize}
\item We identify that the preference pair margin is the key to undermining DPO performance and introduce group reward standardization to alleviate the influence of pair margin.
\item We propose Group Preference Optimization, a self-improvement training framework that breaks the dependence on high-quality data and leverage the inherent ability of the diffusion model to improve various abilities.
\item Extensive quantitative and qualitative comparisons with baseline models indicate that our method can improve the performance in various scenarios, including accurate counting, text rendering, and text-image alignment.
\end{itemize}

\section{Related Works}
\paragraph{Aligning Large Language Models} Aligning LLMs with human preferences\citep{christiano2017deep,bai2022training} has become an inevitable step and de facto standard for improving the performance. RLHF rely on collecting extensive human annotated preference pairs, training reward models to approximate these preferences, and then optimizing LLMs via RL algorithms (e.g., PPO \cite{schulman2017proximal} or REINFORCE\cite{ahmadian2024back}) to maximize reward scores. Different from PPO, which requires a critic model to evaluate policy performance, Group Relative Policy Optimization (GRPO)\cite{shao2024deepseekmath} compares groups of candidate responses directly, eliminating the need for an additional critic model. Recently, Direct Preference Optimization (DPO) \cite{rafailov2023direct} and its variants \cite{ethayarajh2024kto, meng2024simpo,hong2024reference} have emerged as a compelling alternative, offering a mathematically equivalent formulation that bypasses the reward model and optimizes on preference data directly. 

\paragraph{Alignment for Diffusion Models} Motivated by the success of RLHF in LLMs, recent works have introduced several methods for aligning diffusion models. Differentiable reward finetuning approaches \cite{clarkdirectly, wu2024deep} optimize the model directly to maximize the reward of generated images. However, these methods suffer from two key limitations: (1) they require gradient backpropagation through the full denoising chain, resulting in substantial computational overhead; (2) direct access to reward model gradients makes them vulnerable to reward hacking.
DPOK\cite{wu2024deep} and DDPO\cite{blacktraining} formulate the denoising process as a Markov Decision Process (MDP), leveraging reinforcement learning to align diffusion models with specific preferences. 
Diff-DPO\cite{wallace2024diffusion} and D3PO \cite{yang2024using} adapt DPO from language models to diffusion models, achieving superior performance compared to RL-based approaches. Diffusion-KTO\cite{li2024aligning} generalizes the human utility maximization framework to diffusion models, which unlocks the potential of leveraging per-image binary preference signals. While these methods optimize trajectory-level preferences, the preference ordering of intermediate denoising steps may not align with that of the final generated images. Thus, SPO\cite{liang2025aestheticposttrainingdiffusionmodels} proposes a step-aware preference optimization method, which decodes latents at different timesteps to evaluate. LPO\cite{zhang2025diffusion} shares the same idea with SPO, but evaluates on the latent space directly, which can reduce computation overhead. 

\section{Preliminary}
\paragraph{Diffusion Models} Diffusion Models
\cite{ho2020denoising,lipmanflow,liu2023flow} learn to predict data distribution $x_{0} \sim p_{data}(x)$ by reversing the ODE flow. Specifically, with a pre-defined signal-noise schedule $\{\alpha_t, \sigma_t\}_{t=1}^{T}$ on $T$ timesteps, it samples a gaussian noise $\epsilon \sim \mathcal{N}(0, I)$, and constructs a noisy sample $x_t$ at time $t$ as $x_t = \alpha_t x_{0} + \sigma_t \epsilon $. The denoising model $\epsilon_\theta$ parameterized by $\theta$ is trained by minimizing the evidence lower bound (ELBO), and the objective can be simplified to a reconstruction loss:
\begin{equation}
\mathcal{L} = \mathbb{E}_{t\sim[1, T], x_0\sim p(x_0), \epsilon \sim \mathcal{N}(0, 1) }\left [ \left \| \epsilon_\theta(x_t, t, c) - \epsilon \right \|_2^2 \right ] \label{eq:simple-loss}
\end{equation}
where $c$ is the condition information, i.e., image caption. During inference, the model starts from gaussian noise $x_T \sim \mathcal{N}(0, I)$ and iteratively applies the learned noise prediction network $\epsilon_\theta$ to estimate and remove the noise, progressively denoising the latent sample to obtain $x_{t-1}$ at each timestep. The specific form of this denoising process depends on the noise schedule: when $\alpha_t^2 + \sigma_t^2 = 1$, it corresponds to the DDPM, while the condition $\alpha_t + \sigma_t = 1$ characterizes flow-matching. These different scheduling schemes lead to distinct sampling trajectories.

\paragraph{RLHF}
RLHF for diffusion model aims to optimize a conditional distribution $p_\theta(x_0 \mid c)$ such that the reward model $r(c, x_0)$ defined on it is maximized, while regularizing the KL-divergence from a reference model $p_\textrm{ref}$. Specifically, RLHF optimizes a model $p_\theta$ to maximize the following objective:
\begin{equation}
\max_{p_\theta} \mathbb{E}_{c\sim\mathcal{D}_c, x_0\sim p_\theta(x_0 \mid c)}[r(c, x_0)] - \beta \mathbb{D}_{\mathrm{KL}}[p_\theta(x_0\mid c)\parallel p_\theta(x_{\mathrm{ref} } \mid c)]
\label{eq:loss-rlhf}
\end{equation}
where the hyperparameter $\beta$ controls KL-regularization strength.

\paragraph{Diff-DPO} The DPO demonstrate that the following objective is equivalent to the process of explicit reinforcement learning with the reward model $r$:

\begin{equation}
\mathcal{L}_\text{DPO} = 
- \mathbb{E}_{(x^w_0,x^l_0) \sim \mathcal{D}} \log \sigma \left(
\beta \mathbb{E}_{\substack{x^w_{1:T} \sim p_\theta(x^w_{1:T}|x^w_0), \\ x^l_{1:T} \sim p_\theta(x^l_{1:T}|x^l_0)}} \left[ \log \frac{p_{\theta}(x^w_{0:T})}{p_\text{ref}(x^w_{0:T})} -  \log \frac{p_{\theta}(x^l_{0:T})}{p_\text{ref}(x^l_{0:T})} \right]
\right)
\label{eq:loss-dpo-raw}
\end{equation}
However, directly applying \eqrefe{eq:loss-dpo-raw} to diffusion models is not feasible as the log-likelihoods of diffusion models are intractable. Diff-DPO utilizes the evidence lower bound (ELBO), the above loss simplifies to:

\begin{equation}
\mathcal{L}_\text{DPO} = 
- \mathbb{E}_{
(x^w_0, x^l_0, c) \sim \mathcal{D}, t \sim \mathcal{U}(0, T), x^w_t \sim q(x^w_t|x^w_0),x^l_t \sim q(x^l_t|x^l_0)
}
\log \sigma (-\beta (\mathbf{s}(x^w, t, \epsilon)-\mathbf{s}(x^l, t, \epsilon)) )
\label{eq:loss-dpo}
\end{equation}
where $\mathbf{s}(x^*, t, \epsilon) = \|\epsilon -\epsilon_\theta(x_t^*, t)\|^2_2 - \|\epsilon - \epsilon_\text{ref}(x_t^*, t)\|^2_2$. To simplify the expression, the constant $T$ is incorporated into the hyperparameter $\beta$.

\section{Methodology}
\subsection{Pairwise Ranking Undermine DPO}
The DPO objective makes a critical simplifying assumption: all winning samples are equally preferred and all losing samples are equally dispreferred. This formulation ignores the potentially important information contained in the reward margin between pairs. 

\begin{wrapfigure}{r}{0.35\textwidth}
  \centering
  \vspace{-10pt}
  \includegraphics[width=0.35\textwidth]{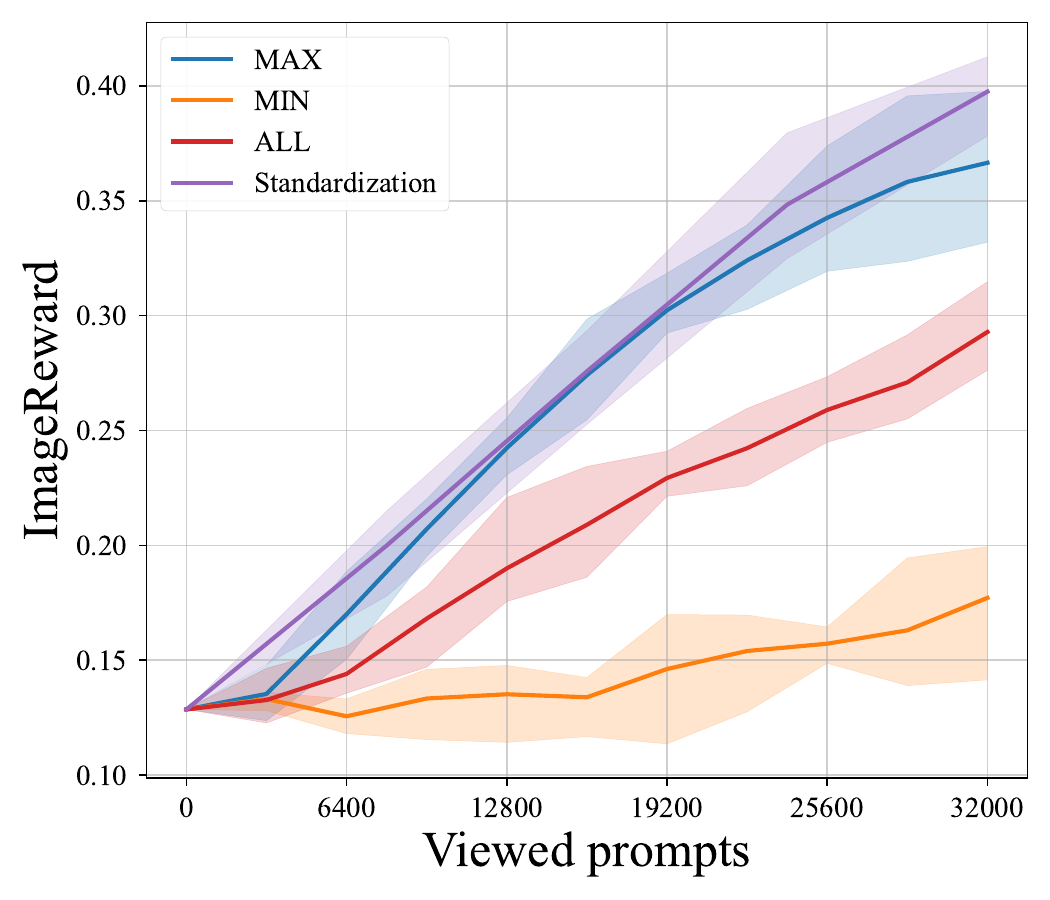}
  \caption{Pair Margin Influence}
  \label{fig:dpo_margin}
  \vspace{-10pt}
\end{wrapfigure}
\paragraph{Problem Hypothesis} 
Suppose we have a reward model $\mathcal{R}$ whose output scores align with human preferences, that is, whenever $\mathcal{R}(x) > \mathcal{R}(y)$, humans prefer $x$ over $y$. While DPO training uses preference pairs $(x^w, x^l)$ that only provide ordinal information ($\mathcal{R}(x^w) > \mathcal{R}(x^l)$), it discards the pair margin $\Delta(x^w, x^l) = |\mathcal{R}(x^w) - \mathcal{R}(x^l)|$. We hypothesize that ignoring this pair margin $\Delta$ leads to suboptimal DPO training. 

\paragraph{Empirical Validation} 
We conduct controlled experiments using ImageReward\cite{xu2023imagereward} as our reward model $\mathcal{R}$. Firstly, generate four distinct images per prompt from different noise. Then, compute reward scores, yielding $C_4^2=6$ possible pairs per prompt. We trained DPO using three pair selection strategies: all pairs(ALL), pairs with the largest margin(MAX), and smallest margin(MIN). As shown in ~\figrefe{fig:dpo_margin}, MAX pairs achieve both faster convergence and superior final performance, while MIN pairs show sluggish improvement. Training on all pairs yields intermediate results. This indicates that high-margin pairs provide stronger learning signals for preference optimization, while low-margin pairs degrade performance.

\paragraph{Intuitive Explanation} 
Consider a reward ranking $A(0.9) \succ B(0.1) \succ C(0.07) \succ D(0.05)$, where  $A$ is superior, and others have nearly indistinguishable scores. Current DPO equally treats clear wins ($(A,B)$) and noise-level differences ($(C,D)$) and misinterprets ambiguous pairs ($(B,C)$, $(B,D)$) as meaningful preferences. This indiscriminate treatment of all pairs can misguide optimization, particularly when reward differences fall below the noise threshold of human annotation. Margin-aware approach is necessary to address this limitation.

\subsection{Group Preference optimization}
Previous works \cite{lee2025calibrated,liu2024videodpo,deng2025less,karthik2024scalable} enhance DPO performance through pair filtering or reward calibration, yet remain confined to pairwise comparisons. In contrast, we propose a groupwise optimization approach that directly leverages reward scores, eliminating the need for pairwise preferences.

\paragraph{Groupwise Formulation} Given a group of $G$ images $\{x^i\}_{i=0}^{G-1}$ ranked by preference (where $x^0 \succ x^1 \succ \cdots \succ x^{G-1}$), we naturally extend pairwise comparisons to all possible $(i,j)$ pairs within the group and formulate the Group DPO loss as $\sum_{0\le i < j < G} - \log \sigma \left( -\beta (\mathbf{s}(x^i, t, \epsilon)-\mathbf{s}(x^j, t, \epsilon)) \right)$. Through algebraic manipulation (see Appendix~\ref{sec:group_dpo}), we derive an equivalent but computationally efficient form $\sum_{i=0}^{G-1} \left[(G-1-2i) \cdot \mathbf{s}(x^i, t, \epsilon)\right]$. This weighting automatically satisfies that higher-ranked items receive a larger positive coefficient, the mean of the group coefficient is zero, and the variance is fixed. Compared to pairwise, groupwise has higher information density and captures $C_G^2$ implicit pairwise comparisons per prompt and reduces $\mathcal{O}(G^2)$ comparisons to $\mathcal{O}(G)$ computation.

\paragraph{Standardization Rewards} Considered the optimization direction in DPO is fundamentally governed by the sign of its coefficient term, we establish groupwise baselines through mean reward scores. Specifically, given a textual prompt $c$ and group of images $\{x^i\}_{i=0}^{G-1}$, a evaluator is used to score the images, yielding $\mathbf{r}=\{r^i\}_{i=0}^{G-1}$ rewards correspondingly. The group of samples is partitioned by mean reward $\overline{\mathbf{r}}=\textbf{mean}(\mathbf{r})$ into winning ( $\mathbf{r}>\overline{\mathbf{r}}$ ) and losing subsets ($\mathbf{r}<\overline{\mathbf{r}}$). Moreover, to ensure scale-invariant optimization steps across varying reward dimensions, we normalize the coefficient by the group reward standard deviation, resulting in stabilized gradient magnitudes. The combined formulation $\frac{\textbf{r} - \textbf{mean}(\textbf{r})}{\textbf{std}(\textbf{r})}$ maintains directional fidelity while adaptively adjusting step sizes based on group score distributions. 

\paragraph{GPO Objective} We replace $(G-1-2i)$ terms in Group DPO with standardized rewards and propose the Group Preference Optimization (GPO) objective, which fine-tunes the model to maximize the rewards of the entire group. The GPO objective is defined as:
\begin{equation}
\mathcal{L}_\text{GPO} = \mathbb{E}_{t\sim\mathcal{S}(1, T), x_0\sim p(x_0), \epsilon \sim \mathcal{N}(0, 1) }\sum_{i=1}^{G} \left [ \mathcal{A}_i (\|\epsilon -\epsilon_\theta(x^i_t, t)\|^2_2 - \|\epsilon - \epsilon_\text{ref}(x^i_t, t)\|^2_2) \right] \label{eq:GPO-loss}
\end{equation}
where $\mathcal{A}_i = \frac{r_i-\mathrm{mean}(\mathbf{r})}{\mathrm{std}(\mathbf{r})}$ is the standardization coefficient, $\mathcal{S}$ is the shifted timestep sampling strategy proposed in SD3\cite{esser2024scaling}. As illustrated in the \figrefe{fig:dpo_margin}, under the same setting, training with GPO loss on all the data not only substantially outperforms Group DPO but also surpasses MAX in performance. 

\paragraph{Efficient Self-Improvement Training} High-quality data for T2I tasks typically relies on images from more powerful generative models, which poses a significant challenge in collecting data. Prior works\cite{zhou2024golden, ahn2024noise, wang2024silent} have observed that certain initial noise conditions can lead to higher-quality images, suggesting that the model inherently possesses the capability to produce superior samples, albeit unstable. Leveraging this insight, we propose a self-improvement framework where the model generates its training samples, bypassing the need for an external model. 
Training with the self-generated data requires sampling from noise to $x_0$, which is computationally intensive. To improve the utilization, we reuse the generated data. Specifically, for each generated data, $k$ timesteps will be randomly sampled at one time for gradient update, and this step will be repeated $\tau$ times. This is a training method that achieves a trade-off between online and offline. The complete pseudo-code of GPO is summarized in Algorithm \ref{alg:gpo-pseudocode}.

\subsection{Design of Reward Score}
Since standardization involves division by the standard deviation, poorly designed reward functions can yield sparse rewards, potentially causing division-by-zero errors. While this issue rarely occurs in tasks with continuous scores (e.g., those using ImageReward), it becomes problematic in tasks like accurate count, where rewards are only given for completely correct responses, resulting in extremely sparse rewards for challenging samples.
Thus, we use a relaxed reward formulation (illustrated in \tabrefe{tab:task-evaluator}). Nevertheless, for edge cases where all rewards are identical, we simply skip that group.
\begin{table}[htbp]\scriptsize
  \caption{Evaluation Score Design}
  \label{tab:task-evaluator}
  \centering
  \begin{tabular}{llll}
    \toprule
    Task & Evaluator & Data Format & Score \\
    \midrule
    \makecell[l]{Accurate \\ Counting} & YOLO & \makecell[l]{Prompt: 2 dogs play with a cat on table \\ Target: [(dog, 2), (cat, 1), (table, 1)]} & \makecell[l]{Single object: $\frac{|N_\text{det}-N_\text{target}|}{N_\text{target}}$ \\ Multi object: average of single case} \\
    \midrule
    \makecell[l]{Text \\ Render} & PPOCR & \makecell[l]{Prompt: A cat hold sign says 'Hello NeurIPS' \\ Target: ('Hello', 'NeurIPS')} & $\text{IoU} = \frac{|S_\text{ocr}\cap S_\text{target}|}{|S_\text{ocr}\cup S_\text{target}|} $ \\
    \midrule
    \makecell[l]{Text \\ Image \\ Align} & BLIP-VQA & \makecell[l]{Prompt: A dog wear sun glass sit on the right \\of a white cat \\ Question(Yes/No): \\ 1. is there a dog? 2. the dog wear a sun glass? \\ 3. is there a cat? 4. is dog on the right of cat?} & The proportion of 'yes' answers i.e. $\frac{N_\text{yes}}{N_\text{total}}$ \\
    \bottomrule
  \end{tabular}
\end{table}

\section{Experiment}
\paragraph{Models} Our method is a general-purpose algorithm compatible with diverse diffusion architectures. We evaluate it on four models: Stable Diffusion 1.5 (SD1.5)\cite{rombach2022high}, Stable Diffusion XL-1.0 Base (SDXL)\cite{podellsdxl}, Stable Diffusion 3.5 Medium (SD3.5M)\cite{esser2024scaling}, and Wan2.1-1.3B(Wan)\cite{wang2025wan}. This selection covers both UNet and DiT backbones, DDPM/flow-matching schedulers, and text encoders ranging from CLIP to T5-XXL.

\paragraph{Datasets and Evaluator} The training data format and evaluation metrics are detailed in ~\tabrefe{tab:task-evaluator}. For each task, we collect 100 prompts from publicly available sources. To enhance prompt diversity efficiently, we utilized the few-shot and instruction-following capabilities of LLMs to generate an additional 1,500 prompts (see Appendix ~\ref{sec:data_build} for details). We reserved 30\% of the data for testing.

\paragraph{Hyperparameter} We perform GPO with group size 32 and fine-tune the model with full parameters by default. We use AdamW\cite{loshchilov2017decoupled} optimizer, and the learning rate is around 2e-8. Further details about the hyperparameter and training are provided in Appendix ~\ref{sec:hyperparameter}.

\subsection{Improvement of Accurate Count and Text Render}
\begin{figure}[htbp]
    \centering
    \begin{tikzpicture}[remember picture]
        \node[inner sep=0pt] (image) {\includegraphics[width=0.95\textwidth]{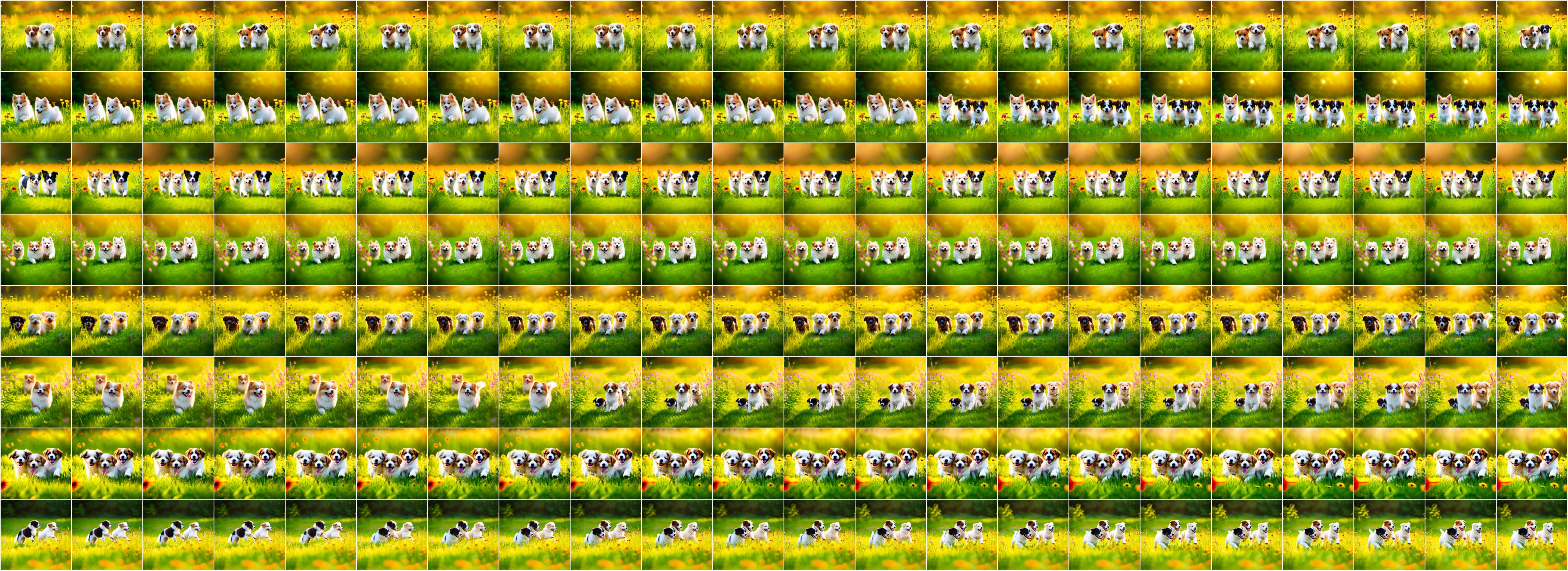}};
        \draw[<->, line width=1pt, >=Stealth]
        ([xshift=-3pt]image.north west) -- ([xshift=-3pt]image.south west)
        node[midway, above=15pt, right=-5pt, rotate=-90] {\textbf{8 Seeds}};
        \draw[->, line width=1pt, >=Stealth]
        ([yshift=5pt]image.north west) -- ([yshift=5pt]image.north east)
        node[midway, above=0pt] {\textbf{Training}};
    \end{tikzpicture}
    \caption{GPO Visualization. Prompt: \textit{There are \textbf{three adorable puppies} playfully running across a lush, sunlit green meadow, their fur glistening in the warm sunlight}}
    \label{fig:training_process_viz}
\end{figure}
\begin{figure}[htbp]
    \centering
    \begin{subfigure}[b]{1.0\textwidth}
        \includegraphics[width=\textwidth]{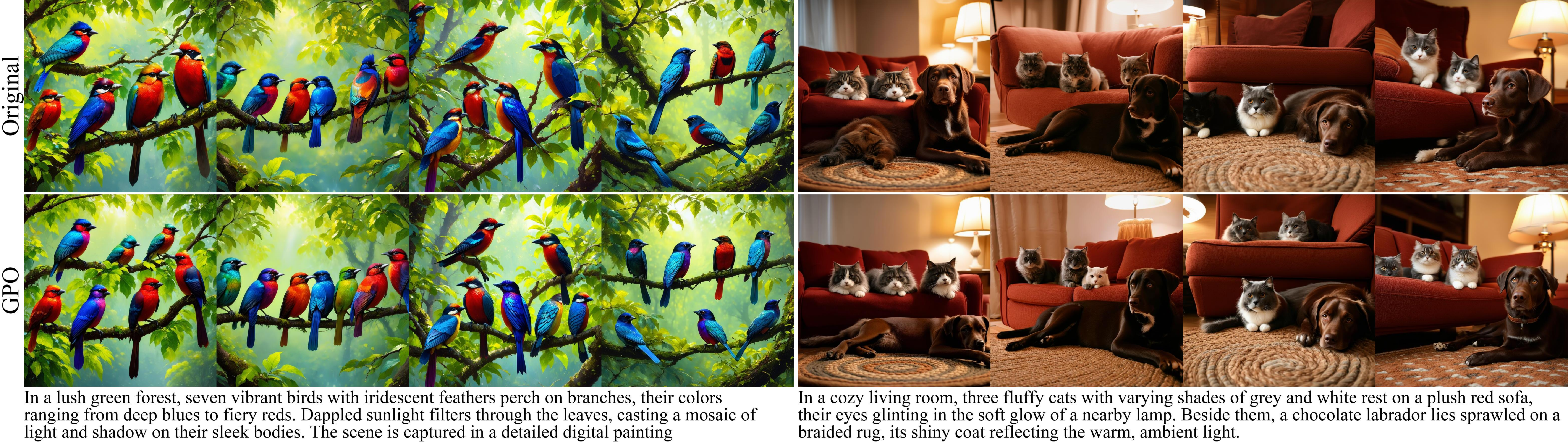}
        \caption{Accurate Count. Left: 7$\times$bird; Right: 3$\times$cat, 1$\times$dog.}
        \label{fig:case_count}
    \end{subfigure}
    \hfill
    \begin{subfigure}[b]{1.0\textwidth}
        \includegraphics[width=\textwidth]{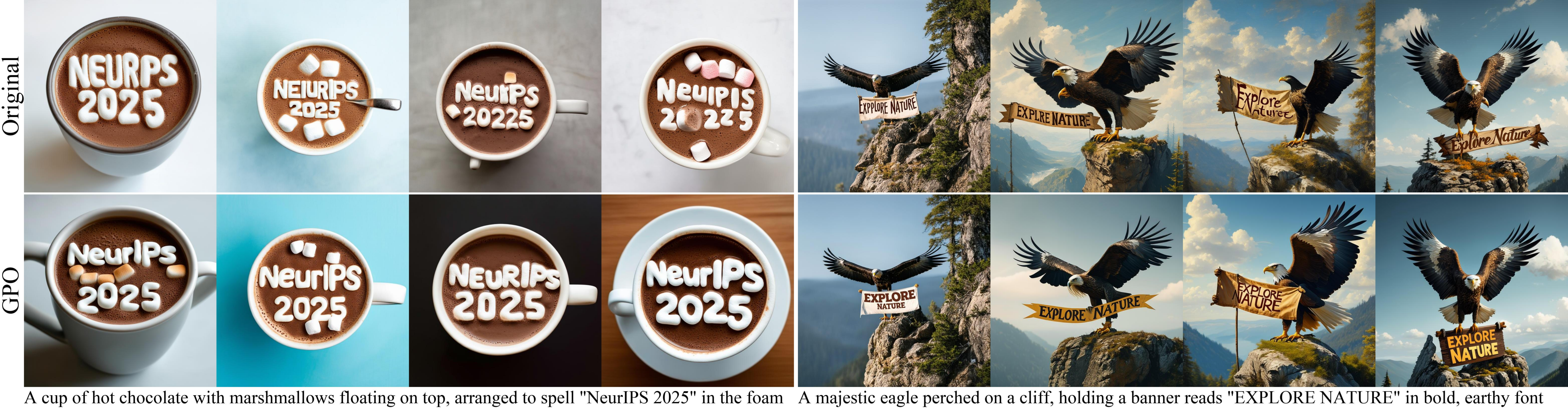}
        \caption{Text Render. Left: "NeurIPS 2025"; Right: "EXPLORE NATURE"}
        \label{fig:case_text}
    \end{subfigure}
    \caption{Qualitative comparisons between SD3.5M and SD3.5M+GPO. All pairs are generated with the same random seed.}
    \label{fig:cases}
\end{figure}

\paragraph{Qualitative result} Through an empirical analysis of samples generated during GPO training, we demonstrate its effectiveness. As shown in \figrefe{fig:training_process_viz}, we generate images using a fixed random seed after each model update. The results exhibit a consistent trend: as training advances, the generated images progressively align with the target prompt, ultimately producing the exact specified count of objects (e.g., 3 dogs). Notably, GPO optimization selectively corrects quantity inaccuracies while preserving the semantic content and structural integrity of the images.
Other qualitative cases can be found \figrefe{fig:cases}.

\paragraph{Quantitative result} Table~\ref{tab:count_text} demonstrates that GPO achieves significant accuracy improvements of approximately 20 percentage points on both tasks for SD3.5M, substantiating the effectiveness of our proposed method. The acurate counting ability of Wan also have similar improvement. However, it fails to enhance the text rendering capability of Wan, as its reliance on self-generated training data inherently limits effectiveness when the base model underperforms.

\begin{table}[htbp]\small
  \setlength{\tabcolsep}{10pt}
  \caption{Quantitative result of accurate counting and text rendering. We also report Pass@4, which evaluates the probability a model can generate the completely correct image out of 4 trials.}
  \label{tab:count_text}
  \centering
  \begin{tabular}{rlllll}
    \toprule
    \multicolumn{1}{c}{\multirow{2}{*}{\bf Model}} & \multicolumn{2}{c}{\bf Accurate Count } & \multicolumn{3}{c}{\bf Text Render} \\
    \cmidrule(lr){2-3}\cmidrule(lr){4-6} & 
    {\bf Accuracy} & {\bf Pass@4} &  {\bf IoU} & {\bf Accuracy} & {\bf Pass@4}\\
    \midrule
    Wan1.3B & 29.3 & 49.7 & 0.024 & 1.1 & 3.2  \\
    \cellcolor{cyan!15}+Ours & \cellcolor{cyan!15}52.2\increase{22.9} & \cellcolor{cyan!15}75.5\increase{25.8} & \cellcolor{cyan!15}0.050\increase{0.026} & \cellcolor{cyan!15}2.3\increase{1.2} & \cellcolor{cyan!15}4.5\increase{1.3} \\
    \midrule
    SD3.5M & 41.8 & 66.5 & 0.258 & 12.8 & 31.9 \\
    \cellcolor{cyan!15}+Ours & \cellcolor{cyan!15}61.1\increase{19.3} & \cellcolor{cyan!15}88.4\increase{21.9} & \cellcolor{cyan!15}0.485\increase{0.227} & \cellcolor{cyan!15}28.1\increase{15.3} & \cellcolor{cyan!15}56.2\increase{24.3} \\
    \bottomrule
  \end{tabular}
\end{table}

\subsection{Evaluation on Compositional Text-Image Alignment}

\begin{table}[htbp]\small
\caption{Quantitative results on T2I-CompBench++\cite{huang2023t2i} and DPG-Bench\cite{hu2024ella}. \textcolor[HTML]{00A86B}{$\uparrow$} and \textcolor[HTML]{E34234}{$\downarrow$} indicate the increase or decrease relative to the original model after GPO.}
\label{tab:t2ialign}
\begin{subtable}{1.0\textwidth}
  \setlength{\tabcolsep}{4pt}
  \centering
  \begin{tabular}{rlllllll}
    \toprule
    \multicolumn{1}{c}{\multirow{2}{*}{\bf Model}} & \multicolumn{3}{c}{\bf Attribute Binding } & \multicolumn{3}{c}{\bf Object Relationship} & \multirow{2}{*}{\bf Complex} \\
    \cmidrule(lr){2-4}\cmidrule(lr){5-7} & 
    {\bf Color} & {\bf Shape} & {\bf Texture} & {\bf 2D-Spatial} & {\bf 3D-Spatial} & {\bf Numeracy} \\
    \midrule
    SD-XL & 52.42 & 44.95 & 50.11 & 17.92 & 31.75 & 47.73 & 34.85 \\
    +DPO & 51.96 & 45.11 & 50.43 & 16.31 & 30.78 & 48.22 & 35.01 \\
    \cellcolor{cyan!15}+Ours & \cellcolor{cyan!15}54.94\increase{3.58} & \cellcolor{cyan!15}47.70\increase{2.75} & \cellcolor{cyan!15}53.94\increase{3.83} & \cellcolor{cyan!15}19.80\increase{1.88} & \cellcolor{cyan!15}34.12\increase{2.73} & \cellcolor{cyan!15}51.43\increase{3.70} & \cellcolor{cyan!15}37.05\increase{2.65} \\
    \midrule
    Wan1.3B & 50.16 & 33.80 & 46.15 & 9.97 & 23.96 & 38.27 & 30.99 \\
    \cellcolor{cyan!15}+Ours & \cellcolor{cyan!15}57.74\increase{7.58} & \cellcolor{cyan!15}38.27\increase{4.47} & \cellcolor{cyan!15}50.79\increase{4.64} & \cellcolor{cyan!15}15.11\increase{5.14} & \cellcolor{cyan!15}29.56\increase{5.60} & \cellcolor{cyan!15}44.22\increase{5.95} & \cellcolor{cyan!15}35.20\increase{4.21} \\
    \midrule
    SD3.5M & 78.94 & 56.72 & 71.45 & 33.99 & 40.20 & 60.93 & 38.39 \\
    \cellcolor{cyan!15}+Ours & \cellcolor{cyan!15}82.11\increase{3.17} & \cellcolor{cyan!15}58.60\increase{1.88} & \cellcolor{cyan!15}73.75\increase{2.30} & \cellcolor{cyan!15}33.58\decrease{0.41} & \cellcolor{cyan!15}41.75\increase{1.55} & \cellcolor{cyan!15}62.18\increase{1.25} & \cellcolor{cyan!15}39.20\increase{0.81} \\
    \bottomrule
  \end{tabular}
  \caption{T2I-CompBench++. The average length of the prompt: 8.7 words}
  \label{tab:t2icompbench}
\end{subtable}

\begin{subtable}{1.0\textwidth}
  \centering
  \setlength{\tabcolsep}{8pt}
  \begin{tabular}{rllllll}
    \toprule
    \bf Model     & \bf Overall & \bf Global & \bf Entity & \bf Attribute & \bf Relation & \bf Other \\
    \midrule
    SD-XL & 72.66 & 79.09 & 80.01 & 80.28 & 81.33 & 80.39 \\
    +DPO & 72.78 & 81.86 & 80.64 & 79.84 & 80.70 & 78.59 \\
    \cellcolor{cyan!15}+Ours & \cellcolor{cyan!15}73.20\increase{0.54} & \cellcolor{cyan!15}78.94\decrease{0.15} & \cellcolor{cyan!15}80.70\increase{0.69} & \cellcolor{cyan!15}80.44\increase{0.16} & \cellcolor{cyan!15}81.57\increase{0.24} & \cellcolor{cyan!15}81.73\increase{1.34} \\
    \midrule
    Wan 1.3B & 80.87 & 87.86 & 88.16 & 88.63 & 87.18 & 88.11 \\
    \cellcolor{cyan!15}+Ours & \cellcolor{cyan!15}83.61\increase{2.74} & \cellcolor{cyan!15}89.90\increase{2.04} & \cellcolor{cyan!15}89.82\increase{1.66} & \cellcolor{cyan!15}89.86\increase{1.23} & \cellcolor{cyan!15}89.29\increase{2.11} & \cellcolor{cyan!15}91.24\increase{3.13} \\

    \midrule
    SD3.5M & 84.25 & 86.59 & 91.49 & 89.64 & 90.23 & 86.64 \\
    \cellcolor{cyan!15}+Ours & \cellcolor{cyan!15}85.33\increase{1.08} & \cellcolor{cyan!15}88.83\increase{2.24} & \cellcolor{cyan!15}91.24\decrease{0.25} & \cellcolor{cyan!15}90.13\increase{0.49} & \cellcolor{cyan!15}92.14\increase{1.91} & \cellcolor{cyan!15}89.41\increase{2.77} \\
    \bottomrule
  \end{tabular}
  \caption{DPG-Bench. The average length of the prompt: 67.1 words}
  \label{tab:dpgbench}
\end{subtable}
\vspace{-0.3cm}
\end{table}

\begin{figure}[htbp]
    \centering
    \includegraphics[width=\textwidth]{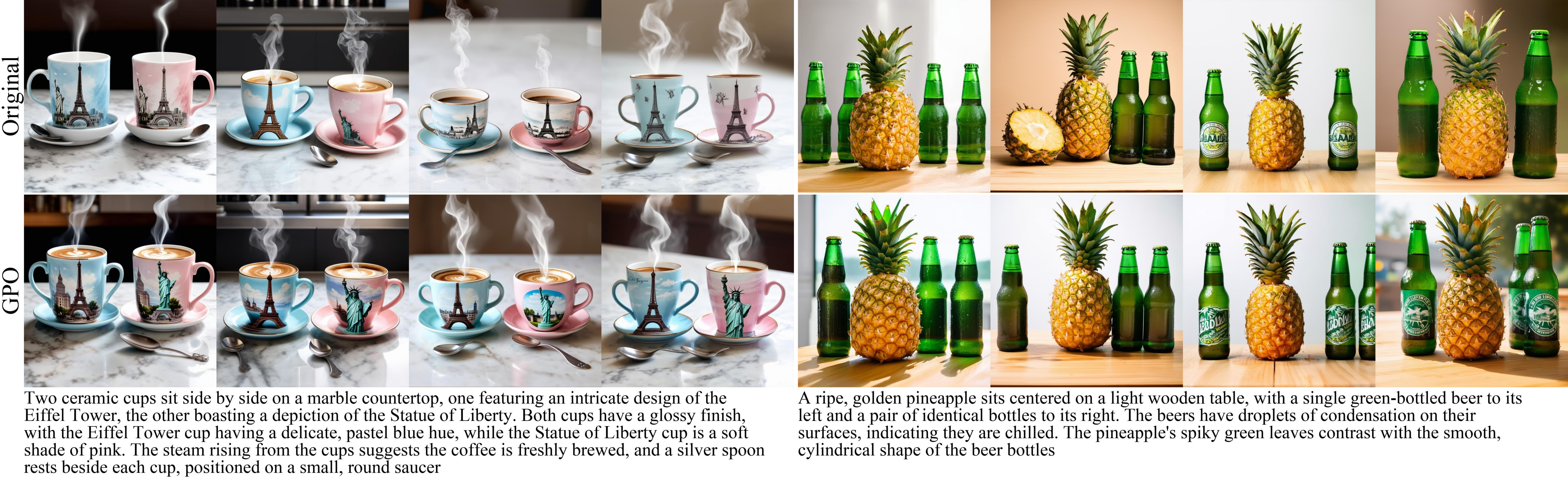}
    \caption{Qualitative comparisons between SD3.5M and SD3.5M+GPO on text-image alignment. All pairs are generated with the same random seed.}
    \label{fig:case_align}
\end{figure}

For quantitative analysis of text-image alignment, we evaluate GPO on two T2I benchmarks: T2ICompbench++\cite{huang2023t2i} for compositional generation and DPG-bench\cite{hu2024ella} for long and detailed prompt understanding. Following official settings, we generate 4 images per prompt for DPG-bench and 10 for T2ICompbench++ to mitigate the influence of randomness.
As shown in Table~\ref{tab:t2ialign}, GPO improves most metrics across models and benchmarks. However, the improvement for SDXL on DPG-Bench is less pronounced, likely due to the benchmark’s focus on evaluating long and complex prompts. Unlike SD3.5M and Wan, which leverage the more capable T5-XXL text encoder, SDXL relies solely on CLIP. Additionally, SD3.5M exhibits smaller gains compared to Wan, as its stronger baseline performance leaves limited room for improvement, and BLIP-VQA may struggle to accurately assess the remaining challenging samples. Qualitative results can be found in \figrefe{fig:case_align}.

\subsection{Comparsion on Aesthetic Preference}
\begin{table}[htbp]\small
\caption{General and aesthetic preference scores on Pick-a-Pic validation unique set except HPS on its benchmark. Comparison methods are evaluated using the official model.}
\label{tab:aes}
\centering
\begin{subtable}{0.48\textwidth}
\centering
\begin{tabular}{lcccc}
\toprule
\bf Method & \bf Aes & \bf P-S & \bf I-R & \bf HPS \\
\midrule
Original & 5.449 & 20.618 & 0.085 & 24.54  \\
Diff-DPO & 5.575 & 21.010 & 0.321 & 25.78  \\
SPO & 5.753 & 21.219 & 0.311 & 27.83  \\
LPO & 5.891 & 21.651 & 0.748 & 28.45  \\
\cellcolor{cyan!15}Ours  & \cellcolor{cyan!15}\textbf{5.951} & \cellcolor{cyan!15}\textbf{21.783} & \cellcolor{cyan!15}\textbf{0.867} & \cellcolor{cyan!15}\textbf{29.11}  \\
\bottomrule
\end{tabular}
\caption{Stable Diffusion 1.5}
\label{tab:sd-aes}
\end{subtable}
% \hfill
\begin{subtable}{0.48\textwidth}
\centering
\begin{tabular}{lcccc}
\toprule
\bf Method & \bf Aes & \bf P-S & \bf I-R & \bf HPS \\
\midrule
Original & 5.971 & 22.094 & 0.802 & 29.31  \\
Diff-DPO & 5.952 & 22.247 & 0.987 & 30.36  \\
SPO & \textbf{6.121} & 22.492 & 1.069 & 31.30  \\
LPO & 6.088 & 22.617 & 1.220 & 31.76  \\
\cellcolor{cyan!15}Ours & \cellcolor{cyan!15}6.115 & \cellcolor{cyan!15}\textbf{22.741} & \cellcolor{cyan!15}\textbf{1.286} & \cellcolor{cyan!15}\textbf{32.25}  \\
\bottomrule
\end{tabular}
\caption{Stable Diffusion XL}
\label{tab:sdxl-aes}
\end{subtable}
\vspace{-0.3cm}
\end{table}

To enable fair comparison with prior work \cite{wallace2024diffusion,yang2024using,zhang2025diffusion,liang2025aestheticposttrainingdiffusionmodels} and demonstrate GPO is also effective under dense reward settings, we evaluate on the aesthetic preference benchmark using both SD-1.5 and SDXL. We quantitatively compare GPO against Diff-DPO \cite{wallace2024diffusion}, SPO \cite{zhang2025diffusion}, and LPO \cite{liang2025aestheticposttrainingdiffusionmodels} using four established metrics: ImageReward (I-R) \cite{xu2023imagereward}, PickScore (P-S) \cite{kirstain2023pick}, Human Preference Score v2.1 (HPS) \cite{wu2023human}, and Aesthetic Score (Aes) \cite{schuhmann2022laion}. Following SPO and LPO, we train GPO using MPS\cite{zhang2024learning} on the 4k prompts from DiffusionDB\cite{wang2022diffusiondb}. ~\tabrefe{tab:aes} reveals two key findings: (1) Alignment methods consistently outperform the vanilla model, with GPO achieving top performance across most metrics, indicating superior human preference alignment; (2) GPO demonstrates strong generalization, showing consistent improvements on metrics not used during training.

\subsection{Further Analysis}
\begin{figure}[htbp]
    \centering
    \begin{subfigure}[b]{0.24\textwidth}
        \includegraphics[width=\textwidth]{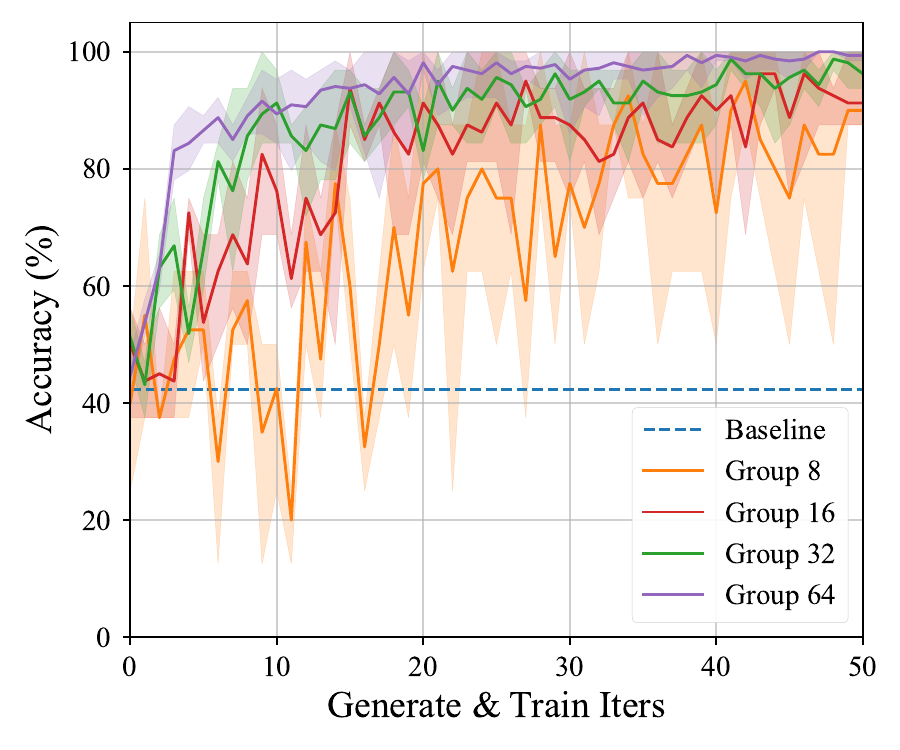}
        \caption{Group Size}
        \label{fig:abla-groupsize}
    \end{subfigure}
    \hfill
    \begin{subfigure}[b]{0.24\textwidth}
        \includegraphics[width=\textwidth]{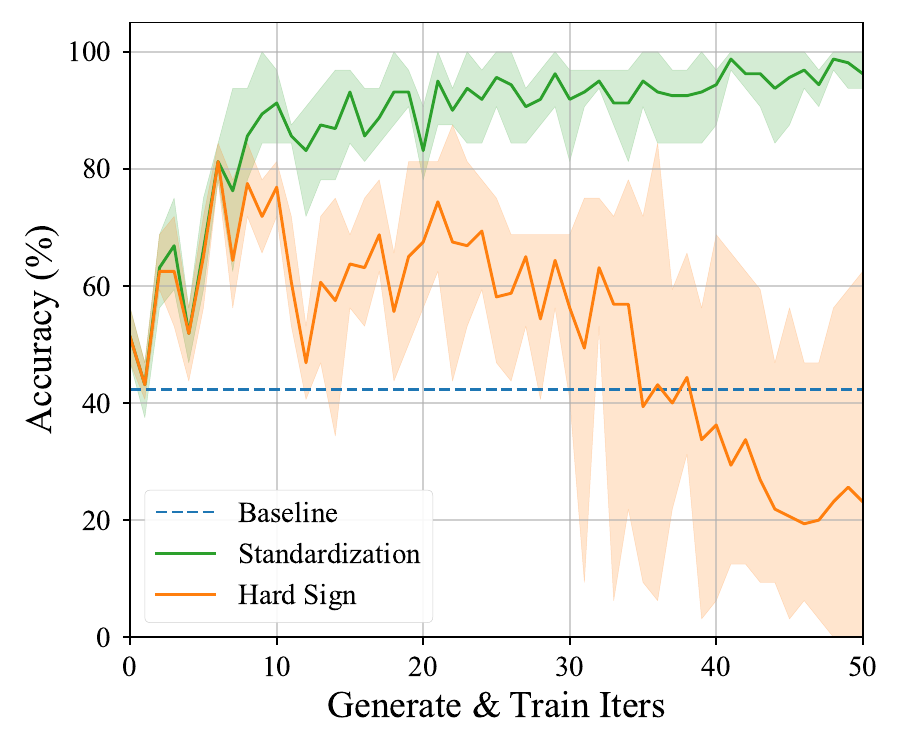}
        \caption{Standardization}
        \label{fig:abla-normalize}
    \end{subfigure}
    \hfill
    \begin{subfigure}[b]{0.24\textwidth}
        \includegraphics[width=\textwidth]{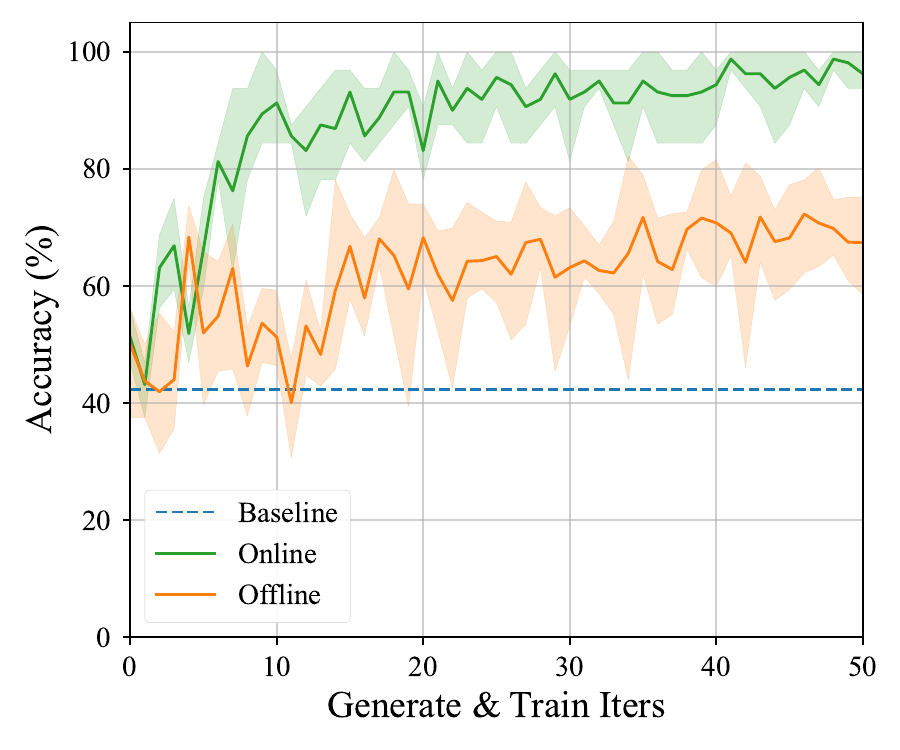}
        \caption{Online/Offline Data}
        \label{fig:abla-online}
    \end{subfigure}
    \hfill
    \begin{subfigure}[b]{0.24\textwidth}
        \includegraphics[width=\textwidth]{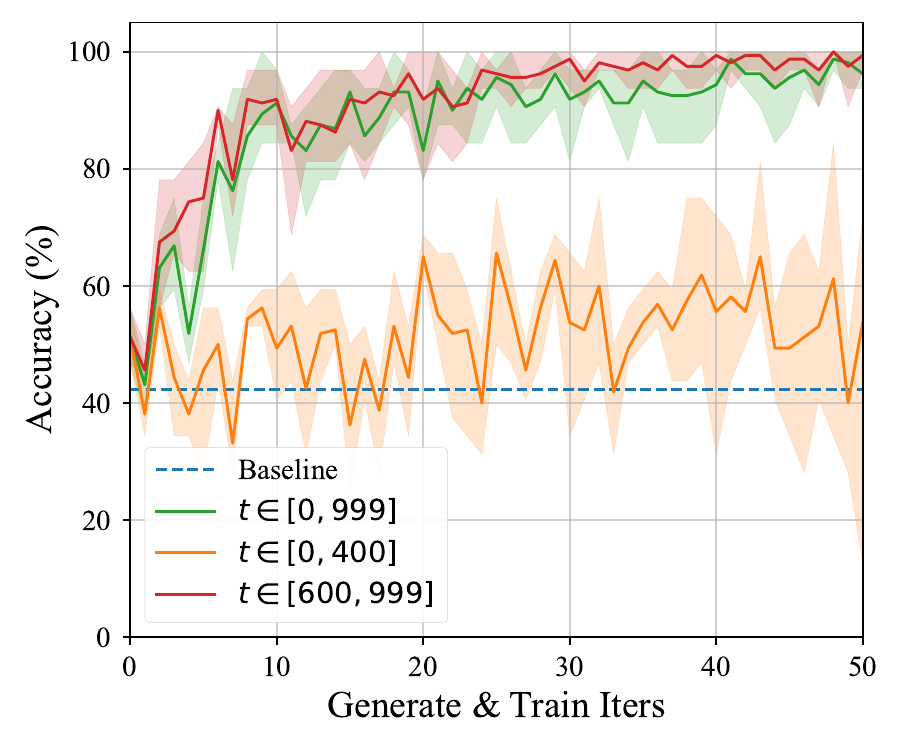}
        \caption{Training Timesteps}
        \label{fig:abla-timesteps}
    \end{subfigure}
    \caption{Alblation Studies of GPO. All experiments are performed on the accurate counting task of Wan-1.3B, repeating each trial with 5 random seeds to ensure robustness.}
    \label{fig:insight}
\end{figure}

\paragraph{Group Size} A key advantage of GPO over prior DPO methods is its use of groupwise comparisons instead of pairwise. We conduct comparative experiments on group sizes $\{8, 16, 32, 64\}$ in \figrefe{fig:abla-groupsize}, showing that larger groups consistently improve training stability and final performance. This stems from richer preference signals, enabling more accurate reward distribution estimation and stable gradient updates. We adopt 32 as the default, as it provides a great trade-off between model performance and computational efficiency.

\paragraph{Standardization} Reward standardization is critical for stable optimization, as it dynamically rebalances sample weights. We conduct an ablation study comparing it against hard sign coefficients \(\mathbf{sgn}(\mathbf{r} - \mathbf{mean}(\mathbf{r}))\), which preserve sign information but discard magnitudes. As shown in \figrefe{fig:abla-normalize}, while both achieve similar initial progress, the hard sign version grows increasingly unstable during training. This instability arises from unnormalized reward variance, e.g., in a group of 32 samples where only one receives a positive reward, optimization becomes dominated by negative gradients. In contrast, standardization stabilizes gradient updates by maintaining consistent coefficient magnitudes.

\paragraph{Online Data} As demonstrated in \figrefe{fig:abla-online}, online data generation consistently outperforms offline approaches, yielding both superior final performance and faster convergence. This improvement arises from the dynamic nature of online generation: as the model advances, the quality of generated samples increases, creating a self-improving feedback loop. In contrast, offline data remains static, ultimately limiting its ability to provide high-quality data during later optimization stages.

\paragraph{Training Timesteps} We compare training strategies focusing exclusively on high-noise, low-noise, and all timesteps. As evidenced by \figrefe{fig:abla-timesteps}, training across all timesteps yields steady performance improvements. In contrast, low-noise-only training results in oscillations around the baseline with marginal gains, while high-noise-only training demonstrates notably more stable convergence. This behavior can be attributed to the inherent properties of diffusion: the low-noise stage primarily refines fine-grained details, leaving higher-level structure (e.g., content and layout) largely unchanged. 

\paragraph{Model Collapse} Since GPO is trained on its own generated data, it inherently suffers from reduced diversity and risks eventual model collapse. Unlike other approaches that use KL regularization, we empirically demonstrate that employing a small learning rate effectively mitigates this issue. 

\subsection{Discussion and Limitations}
Like other self-improvement approaches, GPO is inherently constrained by the capabilities of the base model. If the model lacks a certain ability initially, its own generations may not provide meaningful learning signals for improvement. A promising direction to mitigate this limitation is to bootstrap the desired capability through supervised fine-tuning before applying GPO for iterative refinement. Additionally, GPO incurs higher computational overhead compared to standard fine-tuning, as it requires the diffusion model to perform full inference during training. While this trade-off is justified by the gains in sample quality, future work could explore more efficient data utilization or partial-inference approximations to reduce training costs without sacrificing performance.

\section{Conclusion}
In this paper, we present Group Preference Optimization (GPO), a robust and effective algorithm for self-improvement in T2I models. Our work reveals a critical limitation of DPO: its performance degrades when trained on data pairs with narrow preference margins. To overcome this, we generalize DPO to group-wise comparisons and introduce reward standardization, eliminating the need for pair selection or manual calibration. GPO further reduces dependency on external data by leveraging self-generated samples for training.
Extensive experiments demonstrate that GPO achieves consistent improvements across diverse models and tasks. Notably, by incorporating computer vision models such as YOLO and OCR, our approach enhances fine-grained capabilities like accurate counting and text rendering. These advancements underscore the potential of GPO as a scalable and data-efficient solution for T2I model refinement without requiring external data.

%%%%%%%%%%%%%%%%%%%%%%%%%%%%%%%%%%%%%%%%%%%%%%%%%%%%%%%%%%%%
\newpage
\bibliographystyle{plainnat}
\bibliography{neurips_2025}

\begin{thebibliography}{56}
\providecommand{\natexlab}[1]{#1}
\providecommand{\url}[1]{\texttt{#1}}
\expandafter\ifx\csname urlstyle\endcsname\relax
  \providecommand{\doi}[1]{doi: #1}\else
  \providecommand{\doi}{doi: \begingroup \urlstyle{rm}\Url}\fi

\bibitem[Ahmadian et~al.(2024)Ahmadian, Cremer, Gall{\'e}, Fadaee, Kreutzer, Pietquin, {\"U}st{\"u}n, and Hooker]{ahmadian2024back}
Arash Ahmadian, Chris Cremer, Matthias Gall{\'e}, Marzieh Fadaee, Julia Kreutzer, Olivier Pietquin, Ahmet {\"U}st{\"u}n, and Sara Hooker.
\newblock Back to basics: Revisiting reinforce style optimization for learning from human feedback in llms.
\newblock \emph{arXiv preprint arXiv:2402.14740}, 2024.

\bibitem[Ahn et~al.(2024)Ahn, Kang, Lee, Min, Kim, Jang, Cho, Paul, Kim, Cha, et~al.]{ahn2024noise}
Donghoon Ahn, Jiwon Kang, Sanghyun Lee, Jaewon Min, Minjae Kim, Wooseok Jang, Hyoungwon Cho, Sayak Paul, SeonHwa Kim, Eunju Cha, et~al.
\newblock A noise is worth diffusion guidance.
\newblock \emph{arXiv preprint arXiv:2412.03895}, 2024.

\bibitem[Bai et~al.(2022)Bai, Jones, Ndousse, Askell, Chen, DasSarma, Drain, Fort, Ganguli, Henighan, et~al.]{bai2022training}
Yuntao Bai, Andy Jones, Kamal Ndousse, Amanda Askell, Anna Chen, Nova DasSarma, Dawn Drain, Stanislav Fort, Deep Ganguli, Tom Henighan, et~al.
\newblock Training a helpful and harmless assistant with reinforcement learning from human feedback.
\newblock \emph{arXiv preprint arXiv:2204.05862}, 2022.

\bibitem[Betker et~al.(2023)Betker, Goh, Jing, Brooks, Wang, Li, Ouyang, Zhuang, Lee, Guo, et~al.]{betker2023improving}
James Betker, Gabriel Goh, Li~Jing, Tim Brooks, Jianfeng Wang, Linjie Li, Long Ouyang, Juntang Zhuang, Joyce Lee, Yufei Guo, et~al.
\newblock Improving image generation with better captions.
\newblock \emph{Computer Science. https://cdn. openai. com/papers/dall-e-3. pdf}, 2\penalty0 (3):\penalty0 8, 2023.

\bibitem[Binyamin et~al.(2024)Binyamin, Tewel, Segev, Hirsch, Rassin, and Chechik]{binyamin2024make}
Lital Binyamin, Yoad Tewel, Hilit Segev, Eran Hirsch, Royi Rassin, and Gal Chechik.
\newblock Make it count: Text-to-image generation with an accurate number of objects.
\newblock \emph{arXiv preprint arXiv:2406.10210}, 2024.

\bibitem[Black et~al.(2023)Black, Janner, Du, Kostrikov, and Levine]{blacktraining}
Kevin Black, Michael Janner, Yilun Du, Ilya Kostrikov, and Sergey Levine.
\newblock Training diffusion models with reinforcement learning.
\newblock In \emph{The Twelfth International Conference on Learning Representations}, 2023.

\bibitem[Cao et~al.(2025)Cao, Guo, Huo, Liang, Shi, Song, Zhang, and Zhuang]{cao2025text}
Yuefan Cao, Xuyang Guo, Jiayan Huo, Yingyu Liang, Zhenmei Shi, Zhao Song, Jiahao Zhang, and Zhen Zhuang.
\newblock Text-to-image diffusion models cannot count, and prompt refinement cannot help.
\newblock \emph{arXiv preprint arXiv:2503.06884}, 2025.

\bibitem[Chatterjee et~al.(2024)Chatterjee, Stan, Aflalo, Paul, Ghosh, Gokhale, Schmidt, Hajishirzi, Lal, Baral, et~al.]{chatterjee2024getting}
Agneet Chatterjee, Gabriela Ben~Melech Stan, Estelle Aflalo, Sayak Paul, Dhruba Ghosh, Tejas Gokhale, Ludwig Schmidt, Hannaneh Hajishirzi, Vasudev Lal, Chitta Baral, et~al.
\newblock Getting it right: Improving spatial consistency in text-to-image models.
\newblock In \emph{European Conference on Computer Vision}, pages 204--222. Springer, 2024.

\bibitem[Chen et~al.(2023)Chen, Huang, Lv, Cui, Chen, and Wei]{chen2023textdiffuser}
Jingye Chen, Yupan Huang, Tengchao Lv, Lei Cui, Qifeng Chen, and Furu Wei.
\newblock Textdiffuser: Diffusion models as text painters.
\newblock \emph{Advances in Neural Information Processing Systems}, 36:\penalty0 9353--9387, 2023.

\bibitem[Christiano et~al.(2017)Christiano, Leike, Brown, Martic, Legg, and Amodei]{christiano2017deep}
Paul~F Christiano, Jan Leike, Tom Brown, Miljan Martic, Shane Legg, and Dario Amodei.
\newblock Deep reinforcement learning from human preferences.
\newblock \emph{Advances in neural information processing systems}, 30, 2017.

\bibitem[Clark et~al.(2023)Clark, Vicol, Swersky, and Fleet]{clarkdirectly}
Kevin Clark, Paul Vicol, Kevin Swersky, and David~J Fleet.
\newblock Directly fine-tuning diffusion models on differentiable rewards.
\newblock In \emph{The Twelfth International Conference on Learning Representations}, 2023.

\bibitem[Deng et~al.(2025)Deng, Zhong, Ai, Feng, Wang, and He]{deng2025less}
Xun Deng, Han Zhong, Rui Ai, Fuli Feng, Zheng Wang, and Xiangnan He.
\newblock Less is more: Improving llm alignment via preference data selection.
\newblock \emph{arXiv preprint arXiv:2502.14560}, 2025.

\bibitem[Esser et~al.(2024)Esser, Kulal, Blattmann, Entezari, M{\"u}ller, Saini, Levi, Lorenz, Sauer, Boesel, et~al.]{esser2024scaling}
Patrick Esser, Sumith Kulal, Andreas Blattmann, Rahim Entezari, Jonas M{\"u}ller, Harry Saini, Yam Levi, Dominik Lorenz, Axel Sauer, Frederic Boesel, et~al.
\newblock Scaling rectified flow transformers for high-resolution image synthesis.
\newblock In \emph{Forty-first international conference on machine learning}, 2024.

\bibitem[Ethayarajh et~al.(2024)Ethayarajh, Xu, Muennighoff, Jurafsky, and Kiela]{ethayarajh2024kto}
Kawin Ethayarajh, Winnie Xu, Niklas Muennighoff, Dan Jurafsky, and Douwe Kiela.
\newblock Kto: Model alignment as prospect theoretic optimization.
\newblock \emph{arXiv preprint arXiv:2402.01306}, 2024.

\bibitem[Ho et~al.(2020)Ho, Jain, and Abbeel]{ho2020denoising}
Jonathan Ho, Ajay Jain, and Pieter Abbeel.
\newblock Denoising diffusion probabilistic models.
\newblock In \emph{Proceedings of the 34th International Conference on Neural Information Processing Systems}, pages 6840--6851, 2020.

\bibitem[Hong et~al.(2024)Hong, Lee, and Thorne]{hong2024reference}
Jiwoo Hong, Noah Lee, and James Thorne.
\newblock Reference-free monolithic preference optimization with odds ratio.
\newblock \emph{arXiv e-prints}, pages arXiv--2403, 2024.

\bibitem[Hu et~al.(2024)Hu, Wang, Fang, Fu, Cheng, and Yu]{hu2024ella}
Xiwei Hu, Rui Wang, Yixiao Fang, Bin Fu, Pei Cheng, and Gang Yu.
\newblock Ella: Equip diffusion models with llm for enhanced semantic alignment.
\newblock \emph{arXiv preprint arXiv:2403.05135}, 2024.

\bibitem[Huang et~al.(2023)Huang, Sun, Xie, Li, and Liu]{huang2023t2i}
Kaiyi Huang, Kaiyue Sun, Enze Xie, Zhenguo Li, and Xihui Liu.
\newblock T2i-compbench: A comprehensive benchmark for open-world compositional text-to-image generation.
\newblock \emph{Advances in Neural Information Processing Systems}, 36:\penalty0 78723--78747, 2023.

\bibitem[Jocher et~al.(2023)Jocher, Qiu, and Chaurasia]{YOLO}
Glenn Jocher, Jing Qiu, and Ayush Chaurasia.
\newblock {Ultralytics YOLO}, 2023.
\newblock URL \url{https://github.com/ultralytics/ultralytics}.

\bibitem[Karthik et~al.(2024)Karthik, Coskun, Akata, Tulyakov, Ren, and Kag]{karthik2024scalable}
Shyamgopal Karthik, Huseyin Coskun, Zeynep Akata, Sergey Tulyakov, Jian Ren, and Anil Kag.
\newblock Scalable ranked preference optimization for text-to-image generation.
\newblock \emph{arXiv preprint arXiv:2410.18013}, 2024.

\bibitem[Kirstain et~al.(2023)Kirstain, Polyak, Singer, Matiana, Penna, and Levy]{kirstain2023pick}
Yuval Kirstain, Adam Polyak, Uriel Singer, Shahbuland Matiana, Joe Penna, and Omer Levy.
\newblock Pick-a-pic: An open dataset of user preferences for text-to-image generation.
\newblock \emph{Advances in Neural Information Processing Systems}, 36:\penalty0 36652--36663, 2023.

\bibitem[Labs(2024)]{flux2024}
Black~Forest Labs.
\newblock Flux.
\newblock \url{https://github.com/black-forest-labs/flux}, 2024.

\bibitem[Lee et~al.(2023)Lee, Liu, Ryu, Watkins, Du, Boutilier, Abbeel, Ghavamzadeh, and Gu]{lee2023aligning}
Kimin Lee, Hao Liu, Moonkyung Ryu, Olivia Watkins, Yuqing Du, Craig Boutilier, Pieter Abbeel, Mohammad Ghavamzadeh, and Shixiang~Shane Gu.
\newblock Aligning text-to-image models using human feedback.
\newblock \emph{arXiv preprint arXiv:2302.12192}, 2023.

\bibitem[Lee et~al.(2025)Lee, Li, Wang, He, Ke, Yang, Essa, Shin, Yang, and Li]{lee2025calibrated}
Kyungmin Lee, Xiaohang Li, Qifei Wang, Junfeng He, Junjie Ke, Ming-Hsuan Yang, Irfan Essa, Jinwoo Shin, Feng Yang, and Yinxiao Li.
\newblock Calibrated multi-preference optimization for aligning diffusion models.
\newblock \emph{arXiv preprint arXiv:2502.02588}, 2025.

\bibitem[Li et~al.(2024)Li, Kallidromitis, Gokul, Kato, and Kozuka]{li2024aligning}
Shufan Li, Konstantinos Kallidromitis, Akash Gokul, Yusuke Kato, and Kazuki Kozuka.
\newblock Aligning diffusion models by optimizing human utility.
\newblock In \emph{The Thirty-eighth Annual Conference on Neural Information Processing Systems}, 2024.

\bibitem[Li et~al.(2023)Li, Xu, and Yu]{li2023policy}
Ziniu Li, Tian Xu, and Yang Yu.
\newblock Policy optimization in rlhf: The impact of out-of-preference data.
\newblock \emph{arXiv preprint arXiv:2312.10584}, 2023.

\bibitem[Liang et~al.(2025)Liang, Yuan, Gu, Chen, Hang, Cheng, Li, and Zheng]{liang2025aestheticposttrainingdiffusionmodels}
Zhanhao Liang, Yuhui Yuan, Shuyang Gu, Bohan Chen, Tiankai Hang, Mingxi Cheng, Ji~Li, and Liang Zheng.
\newblock Aesthetic post-training diffusion models from generic preferences with step-by-step preference optimization, 2025.
\newblock URL \url{https://arxiv.org/abs/2406.04314}.

\bibitem[Lipman et~al.(2022)Lipman, Chen, Ben-Hamu, Nickel, and Le]{lipmanflow}
Yaron Lipman, Ricky~TQ Chen, Heli Ben-Hamu, Maximilian Nickel, and Matthew Le.
\newblock Flow matching for generative modeling.
\newblock In \emph{The Eleventh International Conference on Learning Representations}, 2022.

\bibitem[Liu et~al.(2024{\natexlab{a}})Liu, Shao, Li, Bai, Xu, Xiong, Kwok, Helal, and Xie]{liu2024alignment}
Buhua Liu, Shitong Shao, Bao Li, Lichen Bai, Zhiqiang Xu, Haoyi Xiong, James Kwok, Sumi Helal, and Zeke Xie.
\newblock Alignment of diffusion models: Fundamentals, challenges, and future.
\newblock \emph{arXiv preprint arXiv:2409.07253}, 2024{\natexlab{a}}.

\bibitem[Liu et~al.(2022)Liu, Garrette, Saharia, Chan, Roberts, Narang, Blok, Mical, Norouzi, and Constant]{liu2022character}
Rosanne Liu, Dan Garrette, Chitwan Saharia, William Chan, Adam Roberts, Sharan Narang, Irina Blok, RJ~Mical, Mohammad Norouzi, and Noah Constant.
\newblock Character-aware models improve visual text rendering.
\newblock \emph{arXiv preprint arXiv:2212.10562}, 2022.

\bibitem[Liu et~al.(2024{\natexlab{b}})Liu, Wu, Ziqiang, Wei, He, Pi, and Chen]{liu2024videodpo}
Runtao Liu, Haoyu Wu, Zheng Ziqiang, Chen Wei, Yingqing He, Renjie Pi, and Qifeng Chen.
\newblock Videodpo: Omni-preference alignment for video diffusion generation.
\newblock \emph{arXiv preprint arXiv:2412.14167}, 2024{\natexlab{b}}.

\bibitem[Liu et~al.(2023)Liu, Gong, and Liu]{liu2023flow}
Xingchao Liu, Chengyue Gong, and Qiang Liu.
\newblock Flow straight and fast: Learning to generate and transfer data with rectified flow.
\newblock In \emph{The Eleventh International Conference on Learning Representations (ICLR)}, 2023.

\bibitem[Loshchilov and Hutter(2017)]{loshchilov2017decoupled}
Ilya Loshchilov and Frank Hutter.
\newblock Decoupled weight decay regularization.
\newblock \emph{arXiv preprint arXiv:1711.05101}, 2017.

\bibitem[Ma et~al.(2024)Ma, Zong, Song, Li, and Liu]{ma2024exploring}
Bingqi Ma, Zhuofan Zong, Guanglu Song, Hongsheng Li, and Yu~Liu.
\newblock Exploring the role of large language models in prompt encoding for diffusion models.
\newblock In \emph{The Thirty-eighth Annual Conference on Neural Information Processing Systems}, 2024.

\bibitem[Meng et~al.(2024)Meng, Xia, and Chen]{meng2024simpo}
Yu~Meng, Mengzhou Xia, and Danqi Chen.
\newblock Simpo: Simple preference optimization with a reference-free reward.
\newblock \emph{Advances in Neural Information Processing Systems}, 37:\penalty0 124198--124235, 2024.

\bibitem[Podell et~al.(2023)Podell, English, Lacey, Blattmann, Dockhorn, M{\"u}ller, Penna, and Rombach]{podellsdxl}
Dustin Podell, Zion English, Kyle Lacey, Andreas Blattmann, Tim Dockhorn, Jonas M{\"u}ller, Joe Penna, and Robin Rombach.
\newblock Sdxl: Improving latent diffusion models for high-resolution image synthesis.
\newblock In \emph{The Twelfth International Conference on Learning Representations}, 2023.

\bibitem[Rafailov et~al.(2023)Rafailov, Sharma, Mitchell, Manning, Ermon, and Finn]{rafailov2023direct}
Rafael Rafailov, Archit Sharma, Eric Mitchell, Christopher~D Manning, Stefano Ermon, and Chelsea Finn.
\newblock Direct preference optimization: Your language model is secretly a reward model.
\newblock \emph{Advances in Neural Information Processing Systems}, 36:\penalty0 53728--53741, 2023.

\bibitem[Rombach et~al.(2022)Rombach, Blattmann, Lorenz, Esser, and Ommer]{rombach2022high}
Robin Rombach, Andreas Blattmann, Dominik Lorenz, Patrick Esser, and Bj{\"o}rn Ommer.
\newblock High-resolution image synthesis with latent diffusion models.
\newblock In \emph{Proceedings of the IEEE/CVF conference on computer vision and pattern recognition}, pages 10684--10695, 2022.

\bibitem[Schuhmann et~al.(2021)Schuhmann, Vencu, Beaumont, Kaczmarczyk, Mullis, Katta, Coombes, Jitsev, and Komatsuzaki]{schuhmann2021laion}
Christoph Schuhmann, Richard Vencu, Romain Beaumont, Robert Kaczmarczyk, Clayton Mullis, Aarush Katta, Theo Coombes, Jenia Jitsev, and Aran Komatsuzaki.
\newblock Laion-400m: Open dataset of clip-filtered 400 million image-text pairs.
\newblock \emph{arXiv preprint arXiv:2111.02114}, 2021.

\bibitem[Schuhmann et~al.(2022)Schuhmann, Beaumont, Vencu, Gordon, Wightman, Cherti, Coombes, Katta, Mullis, Wortsman, et~al.]{schuhmann2022laion}
Christoph Schuhmann, Romain Beaumont, Richard Vencu, Cade Gordon, Ross Wightman, Mehdi Cherti, Theo Coombes, Aarush Katta, Clayton Mullis, Mitchell Wortsman, et~al.
\newblock Laion-5b: An open large-scale dataset for training next generation image-text models.
\newblock \emph{Advances in neural information processing systems}, 35:\penalty0 25278--25294, 2022.

\bibitem[Schulman et~al.(2017)Schulman, Wolski, Dhariwal, Radford, and Klimov]{schulman2017proximal}
John Schulman, Filip Wolski, Prafulla Dhariwal, Alec Radford, and Oleg Klimov.
\newblock Proximal policy optimization algorithms.
\newblock \emph{arXiv preprint arXiv:1707.06347}, 2017.

\bibitem[Shao et~al.(2024)Shao, Wang, Zhu, Xu, Song, Bi, Zhang, Zhang, Li, Wu, et~al.]{shao2024deepseekmath}
Zhihong Shao, Peiyi Wang, Qihao Zhu, Runxin Xu, Junxiao Song, Xiao Bi, Haowei Zhang, Mingchuan Zhang, YK~Li, Y~Wu, et~al.
\newblock Deepseekmath: Pushing the limits of mathematical reasoning in open language models.
\newblock \emph{arXiv preprint arXiv:2402.03300}, 2024.

\bibitem[Tuo et~al.(2023)Tuo, Xiang, He, Geng, and Xie]{tuo2023anytext}
Yuxiang Tuo, Wangmeng Xiang, Jun-Yan He, Yifeng Geng, and Xuansong Xie.
\newblock Anytext: Multilingual visual text generation and editing.
\newblock In \emph{The Twelfth International Conference on Learning Representations}, 2023.

\bibitem[Wallace et~al.(2024)Wallace, Dang, Rafailov, Zhou, Lou, Purushwalkam, Ermon, Xiong, Joty, and Naik]{wallace2024diffusion}
Bram Wallace, Meihua Dang, Rafael Rafailov, Linqi Zhou, Aaron Lou, Senthil Purushwalkam, Stefano Ermon, Caiming Xiong, Shafiq Joty, and Nikhil Naik.
\newblock Diffusion model alignment using direct preference optimization.
\newblock In \emph{Proceedings of the IEEE/CVF Conference on Computer Vision and Pattern Recognition}, pages 8228--8238, 2024.

\bibitem[Wang et~al.(2025)Wang, Ai, Wen, Mao, Xie, Chen, Yu, Zhao, Yang, Zeng, et~al.]{wang2025wan}
Ang Wang, Baole Ai, Bin Wen, Chaojie Mao, Chen-Wei Xie, Di~Chen, Feiwu Yu, Haiming Zhao, Jianxiao Yang, Jianyuan Zeng, et~al.
\newblock Wan: Open and advanced large-scale video generative models.
\newblock \emph{arXiv preprint arXiv:2503.20314}, 2025.

\bibitem[Wang et~al.(2024)Wang, Huang, Zhu, Russakovsky, and Wu]{wang2024silent}
Ruoyu Wang, Huayang Huang, Ye~Zhu, Olga Russakovsky, and Yu~Wu.
\newblock The silent prompt: Initial noise as implicit guidance for goal-driven image generation.
\newblock \emph{arXiv preprint arXiv:2412.05101}, 2024.

\bibitem[Wang et~al.(2022)Wang, Montoya, Munechika, Yang, Hoover, and Chau]{wang2022diffusiondb}
Zijie~J Wang, Evan Montoya, David Munechika, Haoyang Yang, Benjamin Hoover, and Duen~Horng Chau.
\newblock Diffusiondb: A large-scale prompt gallery dataset for text-to-image generative models.
\newblock \emph{arXiv preprint arXiv:2210.14896}, 2022.

\bibitem[Wu et~al.(2023)Wu, Sun, Zhu, Zhao, and Li]{wu2023human}
Xiaoshi Wu, Keqiang Sun, Feng Zhu, Rui Zhao, and Hongsheng Li.
\newblock Human preference score: Better aligning text-to-image models with human preference.
\newblock In \emph{Proceedings of the IEEE/CVF International Conference on Computer Vision}, pages 2096--2105, 2023.

\bibitem[Wu et~al.(2024)Wu, Hao, Zhang, Sun, Huang, Song, Liu, and Li]{wu2024deep}
Xiaoshi Wu, Yiming Hao, Manyuan Zhang, Keqiang Sun, Zhaoyang Huang, Guanglu Song, Yu~Liu, and Hongsheng Li.
\newblock Deep reward supervisions for tuning text-to-image diffusion models.
\newblock In \emph{European Conference on Computer Vision}, pages 108--124. Springer, 2024.

\bibitem[Xu et~al.(2023)Xu, Liu, Wu, Tong, Li, Ding, Tang, and Dong]{xu2023imagereward}
Jiazheng Xu, Xiao Liu, Yuchen Wu, Yuxuan Tong, Qinkai Li, Ming Ding, Jie Tang, and Yuxiao Dong.
\newblock Imagereward: Learning and evaluating human preferences for text-to-image generation.
\newblock \emph{Advances in Neural Information Processing Systems}, 36:\penalty0 15903--15935, 2023.

\bibitem[Yang et~al.(2024)Yang, Tao, Lyu, Ge, Chen, Shen, Zhu, and Li]{yang2024using}
Kai Yang, Jian Tao, Jiafei Lyu, Chunjiang Ge, Jiaxin Chen, Weihan Shen, Xiaolong Zhu, and Xiu Li.
\newblock Using human feedback to fine-tune diffusion models without any reward model.
\newblock In \emph{Proceedings of the IEEE/CVF Conference on Computer Vision and Pattern Recognition}, pages 8941--8951, 2024.

\bibitem[Zhang et~al.(2023)Zhang, Rao, and Agrawala]{zhang2023adding}
Lvmin Zhang, Anyi Rao, and Maneesh Agrawala.
\newblock Adding conditional control to text-to-image diffusion models.
\newblock In \emph{Proceedings of the IEEE/CVF international conference on computer vision}, pages 3836--3847, 2023.

\bibitem[Zhang et~al.(2024)Zhang, Wang, Wu, Li, Gao, Zhang, and Wang]{zhang2024learning}
Sixian Zhang, Bohan Wang, Junqiang Wu, Yan Li, Tingting Gao, Di~Zhang, and Zhongyuan Wang.
\newblock Learning multi-dimensional human preference for text-to-image generation.
\newblock In \emph{Proceedings of the IEEE/CVF Conference on Computer Vision and Pattern Recognition}, pages 8018--8027, 2024.

\bibitem[Zhang et~al.(2025)Zhang, Da, Ding, Jin, Li, Gao, Zhang, Xiang, and Pan]{zhang2025diffusion}
Tao Zhang, Cheng Da, Kun Ding, Kun Jin, Yan Li, Tingting Gao, Di~Zhang, Shiming Xiang, and Chunhong Pan.
\newblock Diffusion model as a noise-aware latent reward model for step-level preference optimization.
\newblock \emph{arXiv preprint arXiv:2502.01051}, 2025.

\bibitem[Zheng et~al.(2024)Zheng, Teng, Yang, Wang, Chen, Gu, Dong, Ding, and Tang]{zheng2024cogview3}
Wendi Zheng, Jiayan Teng, Zhuoyi Yang, Weihan Wang, Jidong Chen, Xiaotao Gu, Yuxiao Dong, Ming Ding, and Jie Tang.
\newblock Cogview3: Finer and faster text-to-image generation via relay diffusion.
\newblock In \emph{European Conference on Computer Vision}, pages 1--22. Springer, 2024.

\bibitem[Zhou et~al.(2024)Zhou, Shao, Bai, Xu, Han, and Xie]{zhou2024golden}
Zikai Zhou, Shitong Shao, Lichen Bai, Zhiqiang Xu, Bo~Han, and Zeke Xie.
\newblock Golden noise for diffusion models: A learning framework.
\newblock \emph{arXiv preprint arXiv:2411.09502}, 2024.

\end{thebibliography}

%%%%%%%%%%%%%%%%%%%%%%%%%%%%%%%%%%%%%%%%%%%%%%%%%%%%%%%%%%%%
\newpage

\appendix
\section{Group Preferecne Optimization}
\subsection{Group DPO Objective}
\label{sec:group_dpo}
Given a group of $G$ images $\{x^i\}_{i=0}^{G-1}$ ranked by preference (where $x^0 \succ x^1 \succ \cdots \succ x^{G-1}$), we naturally extend pairwise comparisons to all possible $(i,j)$ pairs within the group and can get $\frac{G(G-1)}{2}$ pairs in total.
Considering the monotonicity of $\log \sigma$, we can derive an equivalent but computationally efficient form:
\begin{equation}
\begin{aligned}
\mathcal{L}_\text{Group} &= \sum_{0\le i < j < G} - \log \sigma (-\beta (\mathbf{s}(x^i, t, \epsilon)-\mathbf{s}(x^j, t, \epsilon)) ) \\
&\propto \sum_{0\le i < j < G} (\mathbf{s}(x^i, t, \epsilon)-\mathbf{s}(x^j, t, \epsilon)) =\sum_{i=0}^{G-1}[(G-1-2i) \ \mathbf{s}(x^i, t, \epsilon)]
% \label{eq:Group-loss}
\end{aligned}
\end{equation}
This formula transformation reduces $\mathcal{O}(G^2)$ comparisons to $\mathcal{O}(G)$ computation.

\subsection{Pseudo-code of the GPO}
The complete pseudo-code of gpo is as follows:
\begin{algorithm}
  \small
  \caption{Group Preference Optimization for Diffusion}
  \textbf{Input} reference model $\epsilon_\text{ref}$; evaluator model $\mathcal{R}_\phi$; prompts $\mathcal{D}$; hyperparameters $k$, $\tau$ \\
  \textbf{Output} aligned model $\epsilon_\theta$

  \begin{algorithmic}[1]
    \State policy model $\epsilon_\theta \leftarrow \epsilon_\text{ref}$
    \While{not converged}
        \State Sample batch of prompts $\mathcal{B} \subset  \mathcal{D}$
        \State For prompt $c \in \mathcal{B} $, generate $G$ images $\{x^i\}_{i=1}^G$ from different $x_T$ using $\epsilon_\text{ref}$
        \State Compute rewards $\{r_i\}_{i=1}^{G}$ for generate image $r_i = \mathcal{R}_\phi (x_i)$ 
        \State Compute $\mathcal{A}_{i}$ for the $i$-th image through Standardized operation
        \For{iteration = 1, \dots, $\tau$}
            \State Random sample $k$ timesteps
            \State Update the policy model $\epsilon_\theta$ by minimizing the GPO objective (\eqrefe{eq:GPO-loss})
        \EndFor
        \State Update reference model  $\epsilon_\text{ref} \leftarrow \epsilon_\theta$

    \EndWhile
  \end{algorithmic}
  \label{alg:gpo-pseudocode}
\end{algorithm}

\section{Experiment Details}
\subsection{Model Choosen}
Stable Diffusion 1.5 (SD1.5)\cite{rombach2022high}, Stable Diffusion XL-1.0 Base (SDXL)\cite{podellsdxl}, Stable Diffusion 3.5 Medium (SD3.5M)\cite{esser2024scaling}, and Wan2.1-1.3B(Wan)\cite{wang2025wan} are used in our experiments. This comprehensive selection encompasses diverse architectural paradigms, including both UNet and DiT backbones, and incorporates different training frameworks through DDPM and flow-matching schedulers. The models also employ varying text encoding strategies, ranging from CLIP to the more advanced T5-XXL encoder.
\begin{table}[htbp]
    \centering
    \caption{Hyperparameters of GPO training.}
    \label{tab:exp_detail}
    \begin{tabular}{ccccc}
         \toprule
         & SD 1.5 & SD-XL & SD 3.5 Medium & Wan 2.1 1.3B \\
         \midrule
         Noise Scheduler & DDPM & DDPM & Flow Matching & Flow Matching \\
         Text Encoder & CLIP & CLIP & CLIP, T5XXL & T5XXL \\
         Denoise Backbone & UNet & UNet & MM-DiT & DiT \\
         \midrule
         Prompt Length & 77 & 77 & 77 & 512 \\
         Resolution & 512 & 1024 & 1024 & 480 \\
         Inference Steps & 50 & 50 & 40 & 50 \\
         Guidence Scale & 7.5 & 7.5 & 4.5 & 5.0 \\
         Group Size & \multicolumn{4}{c}{32} \\
         $k$ & \multicolumn{4}{c}{5} \\
         $\tau$ & \multicolumn{4}{c}{3} \\

         \midrule
         Mixed precision & fp16 & fp16 & bf16 & bf16 \\
         Learning Rate & 2e-8 & 2e-8 & 4e-8 & 4e-8 \\
         Optimizer & \multicolumn{4}{c}{AdamW} \\
         Gradient clip & \multicolumn{4}{c}{1.0} \\
         Training Epoch & \multicolumn{4}{c}{2} \\
         GPUs for Training & \multicolumn{4}{c}{8 × NVIDIA A800} \\
         \bottomrule
    \end{tabular}
\end{table}
\subsection{YOLO Detector Choosen}
We employ the widely used YOLOv11 \cite{YOLO} series as our conventional object detector. Benchmark results show that while the extra-large (X) variant offers marginal mAP gains over the large (L) version, its computational latency nearly doubles. Considering the accuracy and computational efficiency of the YOLOv11, we use nano (N), small (S), and large (L) versions during training. For final evaluation, we exclusively use the extra-large (X) model to ensure evaluation robustness and prevent potential metric hacking. However, the YOLO series model, which is trained on the COCO dataset, is unable to detect objects beyond the range of COCO's 80 categories. 

\begin{figure}[h]
    \centering
    \includegraphics[width=0.5 \linewidth]{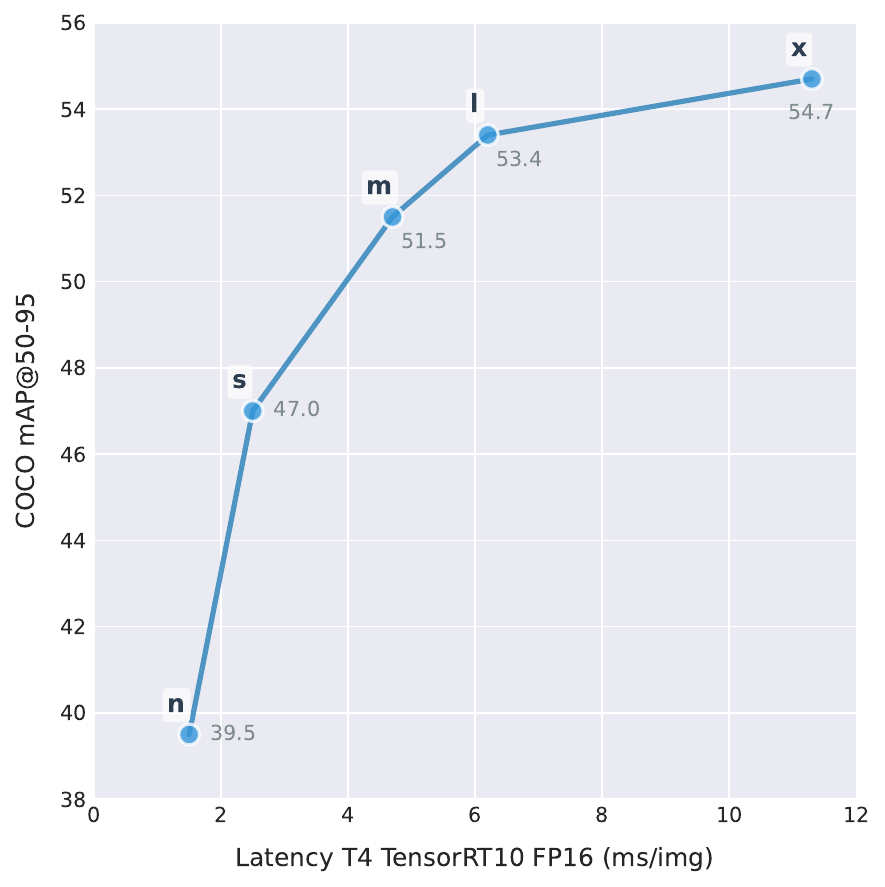}
    \caption{Performance metric of YOLO v11.}
    \label{fig:yolov11_performance_metric}
\end{figure}

\subsection{Hyperparameter Choosen}
\label{sec:hyperparameter}
\paragraph{Group Size} As discussed earlier, we default to using a group size of 32, which achieves a better trade off in terms of performance improvement and training time.

\paragraph{Learning Rate} In our initial verification experiments, we adopted a standard learning rate of 1e-5. However, the model exhibited rapid overfitting, leading to model collapse. Through iterative experimentation, we observed that the training of GPO necessitates an exceptionally small learning rate,on the order of 1e-8. This adjustment not only mitigates overfitting but also enhances model performance.

\paragraph{$k$ and $\tau$ of GPO} These two hyperparameters are designed to enhance the utilization efficiency of the generated data. In our experiments, we empirically set $k=5$ and $\tau=3$ without extensive parameter tuning, as these default values demonstrated satisfactory performance.

\paragraph{Batch Size} Since different models and resolutions require varying amounts of memory, we employ gradient accumulation to maintain a consistent global batch size of $k\cdots G$, thereby ensuring a fixed number of gradient updates. 

\paragraph{Epochs} Since GPO is trained on the online generated data, it achieves notably faste convergence. Remarkably, even a single training epoch yields substantial performance improvements. To strike an optimal balance between model performance and overfitting prevention, we empirically set the default number of training epochs to 2.

\paragraph{Mixed Precision Training} In the experiment, we find that the U-Net architecture exhibits notable sensitivity to numerical precision. To address this, we employ FP16 precision for training the U-Net, while adopting BF16 for the DiT model.

\section{More Qualitative Results}
Since the evaluation of counting and text rendering is relatively objective, we present more examples on these tasks to demonstrate the effectiveness of GPO.

\begin{figure}[htbp]
    \centering
    \includegraphics[width=\textwidth]{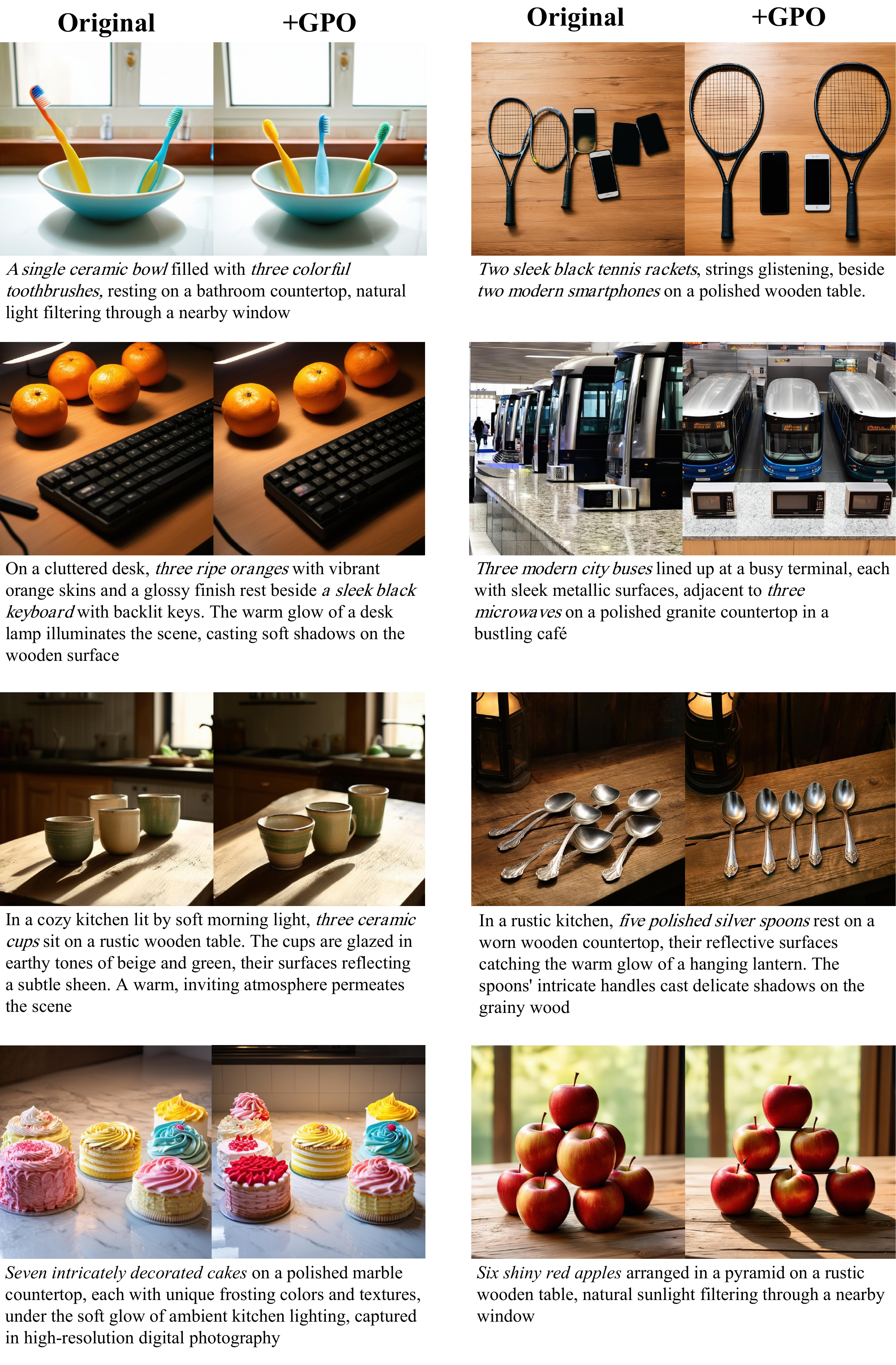}
    \caption{More Comparisons between SD3.5M and SD3.5M+DPO on accurate counting task. All pairs are generated with the same random seed}
\end{figure}

\clearpage
\begin{figure}[htbp]
    \centering
    \includegraphics[width=\textwidth]{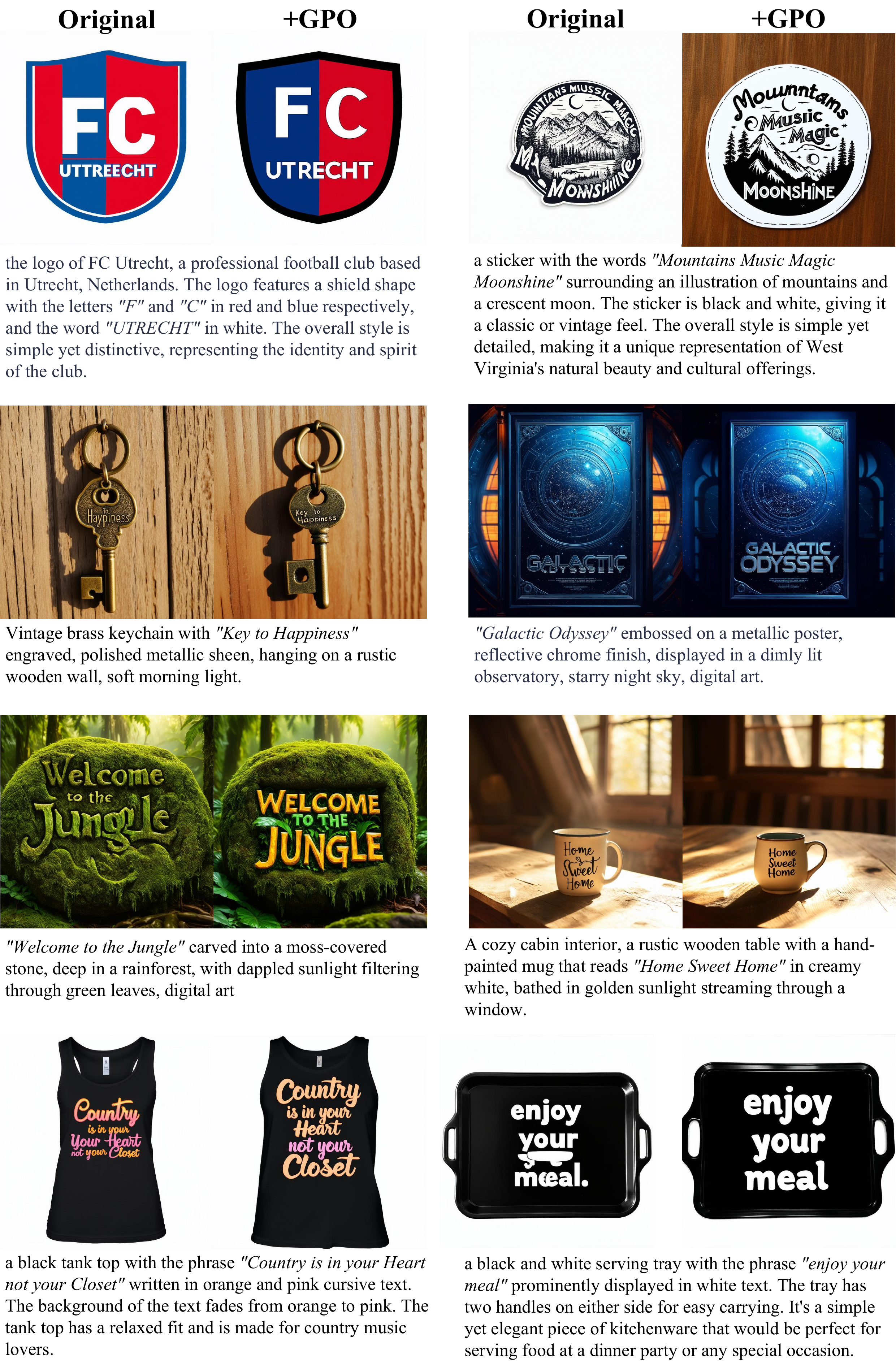}
    \caption{More Comparisons between SD3.5M and SD3.5M+DPO on text rendering task. All pairs are generated with the same random seed}
\end{figure}
\clearpage

\section{Dataset Build Details}
\label{sec:data_build}
To construct high-quality prompts for our experiments, we curated task-relevant prompts from open-source datasets and manually annotated their key components. For each task, we collected an initial set of 100 prompts. During prompt generation, we randomly sampled a subset of these annotated prompts to serve as in-context examples for the large language model (LLM) system prompts. This stochastic selection strategy enhances diversity in the generated outputs. After applying deduplication, we obtained a final dataset of 1,500 unique prompts per task. The system prompt for each task is given below.

% \clearpage
\subsection{Accurate count}
\begin{tcolorbox}[colback=gray!15,colframe=black!40,title=Prompt for Accurate Count Dataset]
[System Instruction] \\
You are a professional prompt engineer specialized in generating high-quality text-to-image captions. Follow these guidelines: \\

[Input Format] \\
User will provide: \\
1. Subject category (e.g., animal/person/scene/object) \\
2. Subject quantity (e.g., single/specific number/plural) \\
3. Optional details (style/action/environment etc.) \\

[Output Requirements] \\
Generate prompts with this structure: \\
1. Core subject: Precise noun phrase \\
2. Visual details: Include color/material/texture \\
3. Environment: Describe setting/lighting/weather \\
4. Art style: Specify photography/painting/digital art etc. \\

[Example Template] \\
Input: 3 cat \\
Output: Three cats curled up together on a sunny windowsill. \\

Input: 4 apple \\
Output: A close-up of 4 fresh green apples with dewdrops, resting on a marble counter. \\

Input: 1 dog, 2 cat \\
Output: A golden retriever sits patiently as two fluffy cats lounge on a cozy living room couch. \\

Input: 2 knife, 2 bowl \\
Output: A simple kitchen scene featuring two knives and two bowls on a marble surface. \\

[Optimization Principles] \\
1. Avoid abstract concepts - use concrete visual elements \\
2. Reduce redundant descriptions \\
3. Separate different dimensions with commas \\
4. Keep under 50 words \\

Generate a prompt for this input: \\
Input: <INPUTS> \\
\end{tcolorbox}
\clearpage

\begin{tcolorbox}[colback=gray!15,colframe=black!40,title=Prompt for Text Render Dataset]
[System Instruction] \\
You are a professional prompt engineer specialized in generating high-quality text-to-image captions. Follow these guidelines: \\

[Output Requirements] \\
Generate prompts with this structure: \\
1. it muse contain text to render wrapped by "" \\
2. Visual details: Include color/material/texture \\
3. optional Environment: Describe setting/lighting/weather \\
4. optional Art style: Specify photography/painting/digital art etc. \\

[Optimization Principles] \\
1. Avoid abstract concepts - use concrete visual elements \\
2. Reduce redundant descriptions \\
3. Separate different dimensions with commas \\
4. Keep under 80 words \\
5. The prompt start should various \\

[Examples] \\
<EXAMPLES> \\

Generate 3 prompts without serial number \\
\end{tcolorbox}

\end{document}